\documentclass{article}

\PassOptionsToPackage{numbers, compress}{natbib}

\usepackage[final]{neurips_2024}

\usepackage[utf8]{inputenc} 
\usepackage[T1]{fontenc}    
\usepackage{hyperref}       
\usepackage{url}            
\usepackage{booktabs}       
\usepackage{amsfonts}       
\usepackage{nicefrac}       
\usepackage{microtype}      
\usepackage{xcolor}         
\usepackage{multirow}
\usepackage{makecell}
\usepackage{array}
\usepackage{graphicx}
\usepackage{enumitem}
\usepackage{amsmath}
\usepackage{rotating}
\usepackage{bbm}
\usepackage{lscape} 
\usepackage[normalem]{ulem}
\usepackage{comment}
\usepackage{bbding}
\usepackage{amssymb}
\usepackage{xspace}
\usepackage{wrapfig}
\usepackage{algorithm}
\usepackage{algpseudocode}
\usepackage{caption}
\usepackage{subcaption}
\usepackage{titletoc}
\newcommand{\myCheckMark}[0]{\textcolor{green}{\Checkmark}}
\newcommand{\myCrossMark}[0]{\textcolor{red}{\XSolidBrush}}
\newcommand{\agentnm}{AGILE\xspace}

\title{\agentnm: A Novel Reinforcement Learning Framework of LLM Agents}

\author{
   \textbf{Peiyuan Feng$^*$}\textsuperscript{1}\quad
   \textbf{Yichen He$^*$}\textsuperscript{1}  \quad
   \textbf{Guanhua Huang$^*$$^\dagger$}\textsuperscript{2} \quad
   \textbf{Yuan Lin$^*$}\textsuperscript{1}\\
   \textbf{Hanchong Zhang$^*$$^\dagger$}\textsuperscript{3}\quad
   \textbf{Yuchen Zhang$^*$}\textsuperscript{1}\quad
   \textbf{Hang Li}\textsuperscript{1}\\
   \textsuperscript{1}ByteDance Research \quad \textsuperscript{2}University of Science and Technology of China\quad \\\textsuperscript{3}Shanghai Jiao Tong University\\
   \texttt{\{fpy,hyc,linyuan.0,zhangyuchen.zyc,lihang.lh\}@bytedance.com,}\\
   \texttt{guanhuahuang@mail.ustc.edu.cn, zhanghanchong@sjtu.edu.cn}
}

\begin{document}

\maketitle

\def\thefootnote{$*$}\footnotetext{Equal contribution. Alphabet order.}

\def\thefootnote{$\dagger$}\footnotetext{Work done during ByteDance Research internship.}

\begin{abstract}
We introduce a novel reinforcement learning framework of LLM agents named \agentnm ({\bf AG}ent that {\bf I}nteracts and {\bf L}earns from {\bf E}nvironments)  designed to perform complex conversational tasks with users, leveraging LLMs, memory, tools, and interactions with experts. The agent possesses capabilities beyond conversation, including reflection, tool usage, and expert consultation. We formulate the construction of such an LLM agent as a reinforcement learning (RL) problem, in which the LLM serves as the policy model. We fine-tune the LLM using labeled data of actions and the PPO algorithm. We focus on question answering and release a dataset for agents called ProductQA, comprising challenging questions in online shopping. Our extensive experiments on ProductQA, MedMCQA and HotPotQA show that AGILE agents based on 7B and 13B LLMs trained with PPO can outperform GPT-4 agents. Our ablation study highlights the indispensability of memory, tools, consultation, reflection, and reinforcement learning in achieving the agent's strong performance. Datasets and code are available at \href{https://github.com/bytarnish/AGILE}{https://github.com/bytarnish/AGILE}.
\end{abstract}

\section{Introduction}

Large Language Models (LLMs) have exhibited remarkable capabilities such as instruction following, reasoning, and zero-shot learning~\cite{brown2020language,wei2022chain,liang2022holistic,openai2023gpt4}, which have greatly catalyzed the development of autonomous agents based on LLMs~\cite{park2023generative,qian2023communicative,bran2023chemcrow}, also known as LLM agents. Recent works propose several essential components or workflows to enhance the abilities of LLM agents, such as planning~\cite{wei2022chain,yao2022react,shen2024hugginggpt}, reflection~\cite{madaan2024self,shinn2024reflexion}, tool-use~\cite{patil2023gorilla,schick2024toolformer,yang2023mm} and life-long learning~\cite{wang2023voyager}. However, it remains unclear how to integrate all components into a unified framework and optimize them end-to-end.

\begin{figure}[thbp]
\vspace{-10pt}
  \centering
   \includegraphics[scale=0.24, trim=50 0 0 0]{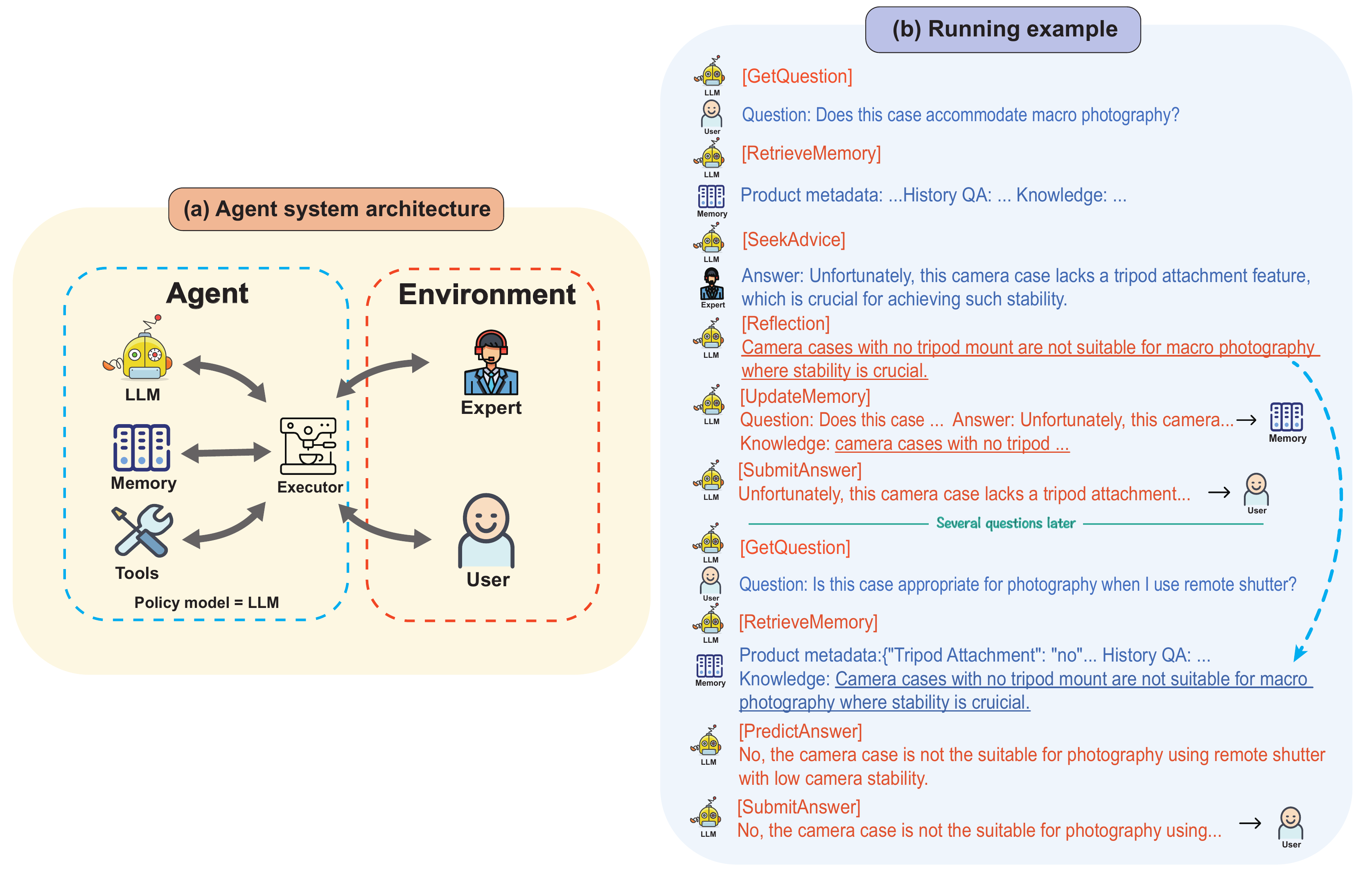}
   \caption{(a) Architecture of our agent system, including LLM, memory, tools, and executor. (b) A running example of \agentnm in a customer service QA environment. The tokens  (actions) generated by the LLM are in orange color and the tokens appended by the executor are in blue color.}
   \vspace{-10pt}
   \label{agent_arch}
\end{figure}

In this paper, we introduce a novel reinforcement learning framework for LLM agents to unify various components and streamline their learning and operation processes. As shown in Figure \ref{agent_arch}(a), the architecture of the agent system, named AGILE, comprises four modules: LLM, memory, tools, and executor. Furthermore, the agent can interact with both users and experts. The LLM, functioning as the predictor of all actions, generates instructions and processes responses. The executor, working as the controller of all actions, interprets the LLM instructions to activate the corresponding modules and collects their responses for the LLM. For example, the executor can fetch a text from the memory and append it to the context of LLM, or extract an excerpt from the context and append it to the memory. The executor can also follow instructions of the LLM to utilize a search tool.
In addition to skills such as reasoning, planning, and reflection, we propose a new ability called \emph{seeking advice}, which means that the agent proactively consults human experts when it encounters a problem unsolvable. The agent can reflect on the expert feedback and memorize it for future use. Furthermore, we propose a training method based on reinforcement learning (RL), which simultaneously trains the policy of invoking different modules and the reasoning, planning, reflection, and seeking advice abilities of the LLM agent in an end-to-end fashion.

While the proposed agent framework is general, in this paper, we evaluate it in complex question answering (QA). It is a task an LLM agent has the potential of outperforming existing solutions such as the use of an LLM alone.  However, existing QA benchmarks~\cite{joshi2017triviaqa,yang2018hotpotqa,jin2021disease,pal2022medmcqa} are designed for specific subsets of capabilities (e.g., reflection, memory retrieve, etc.) which cannot simultaneously investigate the ability to combine all modules and capabilities of the agent.

To address this, we have developed a new benchmark called ProductQA. ProductQA comprises 88,229 question-answer pairs in customer service divided into 26 QA tasks, each corresponding to a distinct Amazon product category. This benchmark is based on real Amazon user queries and includes fact-based questions, reasoning questions, and product recommendation queries. It comprehensively evaluates agents' abilities to handle historical information and accumulated knowledge, leverage tools, interact with humans, perform self-evaluation, and conduct reflection. Additionally, the training and testing tasks are made disjoint to assess the agent's ability to adapt to new product categories.

We evaluate our agent framework on three tasks, ProductQA, MedMCQA~\cite{pal2022medmcqa} and HotPotQA~\cite{yang2018hotpotqa}. For ProductQA, we use a two-stage training method based on Vicuna-13b~\cite{vicuna2023}. In the first stage, imitation learning is employed to create \texttt{agile-vic13b-sft}. In the second stage, the policy gradient algorithm of PPO~\cite{schulman2017proximal} produces \texttt{agile-vic13b-ppo}. Experimental results show that \texttt{agile-vic13b-ppo} improves the relative total performance score by 9.2\% over GPT-4 and by 90.8\% over GPT-3.5. Ablation studies confirm that all modules in Figure~\ref{agent_arch} are indispensable. Specifically, removing tools or memory usage negatively impacts the agent's performance, leading to a 25.9\% or 17.4\% increase in seeking advice, respectively, or a 9.3\% or 4.0\% relative decrease in the total score, respectively. Disabling the seeking advice function results in a 10.7\% decrease in accuracy. Finally, \texttt{agile-vic13b-ppo} achieves a 2.3\% relative increase in total score compared to \texttt{agile-vic13b-sft}, demonstrating the necessity of PPO training. On MedMCQA, we train an \texttt{agile-mek7b-ppo} agent, initialized from Meerkat-7b~\cite{kim2024small}, following the same two-stage procedure. Our agent improves the base LLM's accuracy from 53.4\% to 85.2\% by seeking advice on 31.6\% instances. This accuracy surpasses the SOTA accuracy of 79.1\% by GPT4-MedPrompt~\cite{nori2023generalist}. When all agents are able to seek advice, our agent also outperforms the GPT-4 agent in terms of the total score. For HotPotQA, we use the same two-stage method to train \texttt{agile-vic13b-ppo} from Vicuna-13b. Our agent achieves 67.5\% accuracy, surpassing the strongest baseline of 48.2\%, by seeking advice on 15.6\% of instances. When advice-seeking is enabled for all agents, our agent outperforms GPT-4 by 10.8\% in total score.

The main contributions of this paper are summarized as follows:
\begin{itemize}
\item We propose a novel reinforcement learning framework of LLM agents. It facilitates end-to-end learning of agents. Notably, this framework enables the agent to proactively seek advice from human experts, providing two advantages: 1) It ensures high-level accuracy when dealing with complex and challenging questions, and 2) it fosters learning from humans, thereby enhancing its abilities to adapt to new tasks.
\item We develop a benchmark, ProductQA, to comprehensively evaluate the agent's capabilities in complex question answering.
\item We perform experiments on multiple tasks to verify our framework and show that AGILE agents based on 13B and 7B LLMs trained with PPO can surpass GPT-4 agents.

\end{itemize}

\section{Methods}

\subsection{RL formulation of agent}

Our agent framework comprises four elements: LLM, memory, tools, and executor, see Figure~\ref{agent_arch}(a). The LLM possesses a \emph{context}, defined as the sequence of tokens it utilizes to generate the next token. In RL terminology, the agent conducts a token-level Markov decision process (MDP). The action space $\mathcal{A}$ corresponds to the LLM's vocabulary, with each token representing an action. Hence, the LLM serves as the \emph{policy model}. The agent's state consists of the (context, memory) pair. Upon predicting a new action $a_t$ (i.e., a new token), the LLM transfers control to the executor. The executor applies predefined logic to transition from the current state $s_t$ to the next state $s_{t+1}$, implementing the state transition function $\mathcal{S}\times\mathcal{A}\to \mathcal{S}$ in RL, and then returns control to the LLM to predict the next action. Concurrently, the environment issues a reward $r(s_t,a_t)$.

Let us examine the state transition more closely. For each action, the executor's first operation is to append the token to the context, preparing the LLM for generating the next token. Then, the executor checks a registered list of \emph{functions}. Each function is designed to execute a set of operations, including memory I/O, tool usage, and interaction with the environment. If the action (i.e., the token) matches a function name, the executor will execute the associated function implementation, further mutating the agent state. For instance, if the token is \texttt{[GetQuestion]}, the executor will prompt the user for a new question and append it to the context; if the token is \texttt{[UpdateMemory]}, the executor will write a specific segment of the context into the memory; if the token is \texttt{[ClearContext]}, the executor will reset the context to \texttt{[BOS]}. In summary, the LLM interacts with the memory and tools by predicting function names, relying on the executor to execute these functions. See Table~\ref{agent_function_table} for a full list of functions defined for a QA agent and see Figure~\ref{agent_arch}(b) for a running example.

\begin{table}[ht]
\vspace{-10pt}
\renewcommand{\arraystretch}{1}
\centering
\caption{Functions for an exemplary customer service QA agent. Among them, \texttt{[Reflection]} and \texttt{[PredictAnswer]} are trivial functions, as the executor passes control immediately back to the LLM to start generating result tokens.}
\label{agent_function_table}
\resizebox{0.99\textwidth}{!}{
\begin{tabular}{p{3.2cm}p{10.5cm}}
\toprule[1pt]
\textbf{Function name} & \textbf{Function implementation} \\
\midrule[0.5pt]
\ \texttt{[GetQuestion]}& Prompt the user for a question and append it to the context. \\
\ \texttt{[RetrieveMemory]}& Retrieve relevant entries from the memory and append them to the context.\\
\ \texttt{[SeekAdvice]}& Ask human experts for advice and append it to the context.\\
\ \texttt{[Reflection]}& $\emptyset$\\
\ \texttt{[UpdateMemory]}& Write a specific segment of the context into the memory.\\
\ \texttt{[SearchProduct]}& Extract a search query from the context, then invoke the search tool and append results to the context.\\
\ \texttt{[PredictAnswer]}& $\emptyset$\\
\ \texttt{[SubmitAnswer]}& Extract a predicted answer from the context and submit it to the user.\\
\ \texttt{[ClearContext]}& Reset the context to a single token \texttt{[BOS]}. \\
\bottomrule[1pt]
\end{tabular}
}
\vspace{-0.2cm}
\end{table}

\subsection{Policy learning}

We frame the policy learning problem as a task of training a language model. Consider an agent trajectory $\tau=(s_1,a_1,...,s_n,a_n)$, we derive a \emph{training sequence} denoted as $(e_1,...,e_n)$, where $e_i$ represents the tokens that the executor appends to the context at step $i$. If $a_i$ is a function name token, then $e_i$ is the concatenation of $a_i$ and extra tokens appended by the function execution; otherwise, $e_i=a_i$. In this sequence, $\{a_1,...,a_n\}$ (the first token of each $e_i$) are referred to as action tokens. The LLM context at step $i$, denoted by $c_i$, is a subsequence of the prefix $(e_1,...,e_{i-1})$; $c_i$ may be shorter than $(e_1,...,e_{i-1})$ because the executor can delete context tokens.

In Imitation Learning (IL), we generate trajectories by observing human experts or more proficient agents, then we derive the training sequences to fine-tune the LLM. It is important to point out that (1) the loss is calculated on the action tokens only, and (2) $c_i$ should serve as the attention mask for tokens in $e_i$, as it reflects the true context perceived by the LLM at the time of action prediction. In reinforcement learning (RL), we treat the LLM as the policy model, from which training sequences can be sampled and individual action tokens are assigned rewards. Consequently, the LLM can be optimized using policy gradient methods, such as PPO~\cite{schulman2017proximal}. Analogous to the IL setup, we apply policy gradient updates exclusively to the action tokens and employ $c_i$ as the attention mask.

In some situations, an agent may produce very long trajectories, potentially yielding training sequences that span millions of tokens and are impractical for training. We can leverage the structure of the trajectory to partition it into smaller segments. For instance, if the agent resets its LLM context at the beginning of every QA session, then we can partition by the session boundary. Nevertheless, these sessions are not entirely independent; actions taken in earlier sessions can influence memory, creating lasting effects on subsequent sessions. To tackle this challenge of long-range dependencies, we propose a training algorithm detailed in Appendix~\ref{algorithm}.

\subsection{Interaction with human experts}

Our agent framework enables the agent to proactively seek advice from human experts. For example, the agent can invoke a \texttt{[SeekAdvice]} function to request expert advice. This approach helps in two ways. Firstly, the agent can request the correct answer when its confidence is low, ensuring sufficient accuracy for the application. Secondly, the agent can use \texttt{[Reflection]} to distill general knowledge from the expert advice before storing it in memory. This accumulation of knowledge allows the agent to adapt to new tasks that it has not encountered during training.

Seeking advice involves complex decision-making. The agent must estimate its own confidence in the current session, predict the potential value of the advice for future sessions, and consider the cost of human resources. The optimal trade-off is difficult to annotate manually but aligns well with our RL framework. Specifically, the present risk, future value, and cost of action can all be represented as RL rewards, allowing this skill to be trained as part of the policy model on an end-to-end basis.

\section{The ProductQA dataset}\label{sec:The ProductQA dataset}

We believe that product question answering in a real online shopping environment offers a comprehensive challenge for evaluating LLM agents. First, it demands expert knowledge about millions of products, including their technical specifications, usage in particular scenarios, and compatibility with other products. Second, answering some questions requires the use of tools, such as a product search tool. Third, the continuous emergence of new products necessitates the adaptability of the agent. This has motivated the creation of the ProductQA dataset. Unlike existing online shopping QA datasets~\cite{shen2023xpqa,deng2023product}, which primarily focus on questions about product metadata or page information, ProductQA features more complex queries involving reasoning, expert knowledge, and tool usage (e.g., SQL), providing a comprehensive assessment of an agent's capabilities.

The ProductQA dataset consists of 26 QA tasks, each representing a distinct group of products within a specific category. Each group encompasses 17-20 products. We collected 20 groups for training and 6 for testing, allowing for assessing the agent's adaptability to new tasks. We collected an average of 3,393 question-answer pairs for each product group. The questions within the same group are correlated, as knowledge from one answer may aid in addressing other questions. The dataset statistics are presented in Table~\ref{data_statistic}. 

The dataset is annotated by 20 professional annotators, each with at least a college degree, employed by a commercial data annotation company. We pay the company at market rates for professional annotation. See annotation guidelines in Appendix~\ref{annotation_guidelines}. In addition, we will release the code for the data pre-processing before human annotation. 

\subsection{Product collection}

We gather products from the Amazon Review Data~\cite{ni2019justifying}, which includes product metadata as well as reviews. We initially filter the Amazon Review Data to retain only popular products with at least 100 reviews, then cluster them by category tags. From these clusters, we select 26 based on the size of the cluster, each defined as a \emph{product group}. Subsequently, we sample products from each product group. See Appendix~\ref{product_collection} for more details about product group and product selection.

After the products are collected, annotators compile an information table for each product group. An example of such a table is presented in Table~\ref{headphones_table}. To enhance the efficiency of the annotation process, we employ GPT-4 to extract as many product features as possible from the reviews. These features, together with the product metadata, are provided to the annotators for table creation.

\begin{table}[t]
\centering
\caption{An example of an information table for the headphones group.}
\scalebox{0.74}{
\begin{tabular}{m{1.9cm}m{2.6cm}m{0.9cm}m{1.5cm}m{1.5cm}m{1.7cm}m{1.5cm}m{1.6cm}m{0.5cm}}
\toprule[1pt]
\multicolumn{1}{c}{\multirow{2}{*}[-0.2ex]{\textbf{Product ID}}} & \multicolumn{1}{c}{\multirow{2}{*}[-0.2ex]{\textbf{Title}}} & \multicolumn{1}{c}{\multirow{2}{*}[-0.2ex]{\textbf{Price}}} & \multicolumn{1}{c}{\multirow{2}{*}[-0.2ex]{\textbf{Brand}}} & \multicolumn{1}{c}{\textbf{Headphone}} & \multicolumn{1}{c}{\multirow{2}{*}[-0.2ex]{\textbf{Cable Type}}} & \multicolumn{1}{c}{\textbf{Audio}} & \multicolumn{1}{c}{\textbf{Audio}} & \multicolumn{1}{c}{\multirow{2}{*}[-0.2ex]{\textbf{...}}} \\
&&&&\multicolumn{1}{c}{\textbf{Type}}&&\multicolumn{1}{c}{\textbf{Transmission}}&\multicolumn{1}{c}{\textbf{Output Mode}} \\
\midrule[0.5pt]
B00WSLZFTK & Sennheiser RS 170 & \$11.03 & Sennheiser & over-ear & bluetooth & kleer & stereo & \multicolumn{1}{c}{...} \\
\midrule[0.5pt]
B003AIL2HE & JVC HAEB75B & \$9.99 & JVC & earbud & 3.5mm Jack & analog & bass boost & \multicolumn{1}{c}{...} \\
\midrule[0.5pt]
B01C22IJV0 & Phaiser BHS-530 & \$6.04 & Phaiser & earbud & bluetooth & bluetooth & stereo & \multicolumn{1}{c}{...} \\
\midrule[0.5pt]
B0013OWPV4 & JVC HARX700 & \$2.00 & JVC & over-ear & 3.5mm Jack & analog & stereo & \multicolumn{1}{c}{...} \\
\midrule[0.5pt]
\multicolumn{1}{c}{...} & \multicolumn{1}{c}{...} & \multicolumn{1}{c}{...} & \multicolumn{1}{c}{...} & \multicolumn{1}{c}{...} & \multicolumn{1}{c}{...} & \multicolumn{1}{c}{...} & \multicolumn{1}{c}{...} & \multicolumn{1}{c}{...} \\
\bottomrule[1pt]
\end{tabular}
}
\label{headphones_table}
\vspace{-5pt}
\end{table}

\begin{table}[t]
    \centering
    \caption{Examples of Fact-QA, Search-QA and Reasoning-QA in ProductQA.}
    \scalebox{0.8}{
        \begin{tabular}{m{2.2cm}m{4.5cm}m{6.9cm}m{1.9cm}}
\toprule[1pt]
            \multicolumn{1}{l}{\textbf{Type}} & \multicolumn{1}{l}{\textbf{Question}} & \multicolumn{1}{l}{\textbf{Long Answer}} & \multicolumn{1}{l}{\textbf{Short Answer}} \\
\midrule[0.5pt]

            Fact-QA & What is the size of the neodymium driver used in the JVC HA-EB75 headphones? & The JVC HA-EB75 headphones contain a 13.5 mm neodymium driver in each earpiece, which contributes to the enhanced sound quality. & 13.5 mm \\
\midrule[0.5pt]

            Search-QA & I'm an audiophile always on the move, so I need my music non-stop. Tell me, what's the headphone with the longest playtime you have, either on-ear or in-ear? & I found a product that matches your criteria. `ABCShopUSA Wireless Earbuds True' with asin: B00LJT2EPK & B00LJT2EPK \\
\midrule[0.5pt]

            Reasoning-QA & Will these headphones deliver comparable sound quality to wired alternatives when I am editing music? & No, these headphones may not suit your needs for music editing since they are wireless and can introduce audio compression and slight latency. Such issues can impact the precise listening experience crucial for professional audio editing tasks. & no \\
\bottomrule[1pt]
        \end{tabular}
    }
    \vspace{-10pt}
    \label{qa_examples}
\end{table}

\subsection{QA collection}

We identify three predominant types of questions in online shopping contexts: 1) {\bf Fact-QA}: questions concerning specific product details; 2) {\bf Search-QA}: searches for product recommendations tailored to user preferences; 3) {\bf Reasoning-QA}: questions whose answers require domain-specific reasoning, such as the implications of a product feature. Accordingly, we annotate question-answer pairs for these types. Each question is annotated with both a detailed paragraph-long answer and a concise short answer. The long answer should resemble a response from human customer service, while the short answer consists of a few words.  We train the model to predict both answer types. The accuracy of the long answers is evaluated using GPT-4 (see Appendix~\ref{sec:Prompt templates} for the prompt); the short answers are assessed by exact match and are used for defining rewards for RL training.

\paragraph{Fact-QA} Fact-QAs are constructed from product reviews. For each product, we provide GPT-4 with a batch of 30 reviews, prompting it to generate 20 questions and their corresponding answers before moving on to the next batch. We encourage GPT-4 to create diverse questions. The results are then given to annotators to refine and finalize the question-answer pairs.

\paragraph{Search-QA} Starting with an information table for a given product group, we generate random SQL expressions using a set of predefined rules. These expressions are then translated into natural language questions by GPT-4. The answers are obtained by executing the SQL queries. Subsequently, human annotators thoroughly revise the QA pairs.

\paragraph{Reasoning-QA} As the first step, we collect professional knowledge for each product group. To enhance efficiency, we utilize GPT-4 to generate candidate knowledge entries based on the technical specifications from the information table. These entries are then curated and refined by human annotators. Here is an example of a knowledge entry: \emph{Motherboards with the ATX form factor are ideally suited for high-performance computing tasks and gaming, due to their ample expansion slots for graphics cards and other peripherals that boost computing capabilities.} Finally, annotators develop question-answer pairs from these knowledge entries.

\section{Experiments}\label{sec:Experiments}

\subsection{Experimental setting}\label{sec:exp-setting}

\paragraph{Dataset} We evaluate our agent on three complex QA tasks: ProductQA, MedMCQA and HotPotQA. MedMCQA~\cite{pal2022medmcqa} is a dataset for multiple-choice QA. It consists of questions from medical school entrance examinations. HotPotQA~\cite{yang2018hotpotqa} features natural, multi-hop questions, which challenge an agent's ability to perform reasoning and utilize search tools. For both MedMCQA and HotPotQA, we report results on their respective full dev sets.

\paragraph{Agent definition} Our agent can invoke functions defined in Table~\ref{agent_function_table}. In a typical workflow, the agent prompts the user for a new question at the session start. It can then retrieve memory to get relevant information. The memory can be initialized as empty (ProdcutQA) or with domain knowledge (QA pairs from MedMCQA training dataset). The agent has the option to use external tools, such as product search in ProductQA and article search in HotPotQA), to gather more information. At last, the agent decides whether to predict an answer directly or seek human advice. If the agent seeks advice, it obtains a human answer (ground-truth answer in our setting). The agent can then optionally use a reflection round to extract general knowledge from the human answer, writing both the human answer and the reflected knowledge to its memory. Finally, the agent submits an answer to the user. In our setting, submitting a correct answer incurs a $+1$ reward, while submitting a wrong answer incurs a $0$ reward. Seeking human advice has a fixed $-c$ reward, where $c$ represents \emph{seeking advice cost}. Assuming the human advice always contains a correct answer, then the possible total rewards are $\{0,1,1-c\}$.

\paragraph{Training} The training consists of two stages. First, we construct trajectories from the training data and employ imitation learning to train the agent. Then we apply Algorithm~\ref{alg:session-level-optimization} for further optimization by reinforcement learning. See Appendix~\ref{sec:product_qa_implementation_details} for implementation details. For ProductQA and HotPotQA, the agent's LLM is initialized from Vicuna-13b-1.5. For MedMCQA, we use Meerkat-7b~\cite{kim2024small}, a medical LLM trained with high-quality CoT reasoning paths from 18 medical textbooks and diverse instruction-following datasets. We fine-tune the model for 2 epochs with a learning rate of 1e-5 and a batch size of 64. We implement PPO for 1 epoch with a learning rate of 1e-6 and a batch size of 64. The training runs on NVIDIA-H800. Training times and the number of GPUs for each experiment are reported in Table~\ref{training_times}. The LLM is fully trained without using LoRA.

\paragraph{Evaluation and baselines} We report three metrics for the agent: (a) Advice rate: the rate of seeking human advice; (b) Accuracy: the rate of predicting the correct answer; (c) Total score: the average reward across all sessions, taking the advice rate and the accuracy both into account.

We compare our agent against two types of baselines: 1) Prompting GPT-3.5 (gpt-3.5-turbo-0301) and GPT-4 (gpt-4-0613)~\cite{openai2023gpt4} to directly answer the question, without working in an agent manner, noted as \texttt{gpt3.5-prompt} and \texttt{gpt4-prompt}. 2) Prompting GPT-3.5 and GPT-4 within the AGILE framework, noted as \texttt{agile-gpt3.5-prompt} and \texttt{agile-gpt4-prompt}. We carefully designed prompts for all baselines and they are shown in Appendix~\ref{sec:Prompt templates}.

\begin{table}[t]
\renewcommand{\arraystretch}{1}
\centering
\caption{Results on {ProductQA}. Here, \texttt{X-prompt} represents directly prompting model X; \texttt{agile-X-Y} incorporates model X within the AGILE framework, while Y represents prompting or PPO training. We report results on short and long answers, respectively.  The seeking advice cost is $c=0.3$. Results are averaged over six test tasks. See Table~\ref{overall_result_appendix} for individual product group performance.}
\label{overall_result}
\scalebox{0.9}{
\begin{tabular}{lccccc}
\toprule[1pt]
\multicolumn{1}{l}{\multirow{2}{*}[-0.9ex]{\textbf{Method}}} & \multirow{2}{*}[-0.9ex]{\makecell[c]{\textbf{Advice}\\\textbf{Rate}}~$\downarrow$ } & \multicolumn{2}{c}{\textbf{Accuracy}~$\uparrow$} & \multicolumn{2}{c}{\textbf{Total Score}~$\uparrow$} \\
\cmidrule[0.5pt](r){3-4} \cmidrule[0.5pt](r){5-6}
& & \textbf{Short} & \textbf{Long} & \textbf{Short} & \textbf{Long} \\
\midrule[0.5pt]
gpt3.5-prompt & - & 0.202 & 0.322 & - & - \\
gpt4-prompt & - & 0.464 & 0.571 & - & - \\
agile-vic13b-prompt & 0.174 & 0.174 & 0.294 & 0.122 & 0.242 \\
agile-gpt3.5-prompt & 0.323 & 0.508 & 0.644 & 0.411 & 0.547 \\
agile-gpt4-prompt & 0.208 & 0.780 & 0.809 & 0.718 & 0.747 \\
\midrule[0.5pt]
agile-vic7b-ppo (ours) & 0.179 & 0.818 & 0.800 & 0.764 & 0.746 \\
agile-vic13b-ppo (ours) & 0.233 & \textbf{0.854} & \textbf{0.854} & \textbf{0.784} & \textbf{0.784} \\
\bottomrule[1pt]
\end{tabular}
}
\vspace{-10pt}
\end{table}

\subsection{Results on ProductQA} \label{sec:Results}

\begin{figure}[htbp]
\centering
\begin{minipage}[t]{0.45\textwidth}
\centering
\includegraphics[width = 1.0 \textwidth]{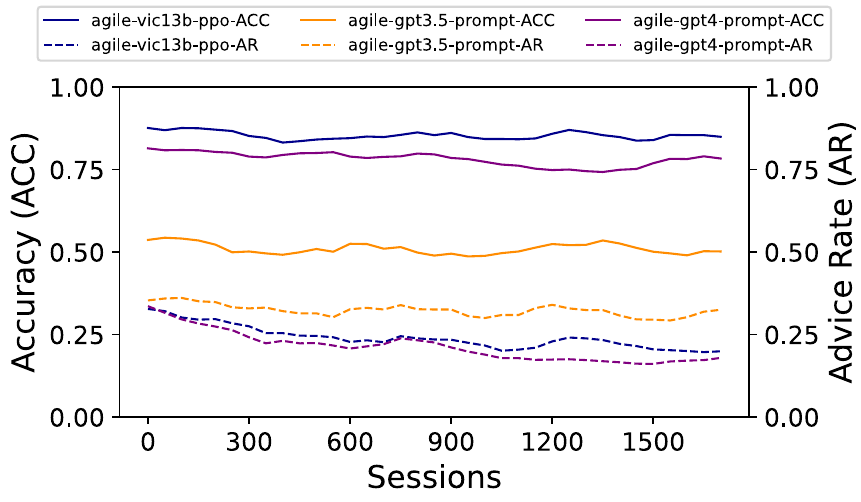}
\vspace{-0.5cm}
\caption{Accuracy and advice rate over the following 200 sessions ($c = 0.3$).
}
\label{fig:acc_sa_comparison}
\end{minipage}
\hspace{.1in}
\begin{minipage}[t]{0.45\textwidth}
\centering
\includegraphics[width=1.0\textwidth]{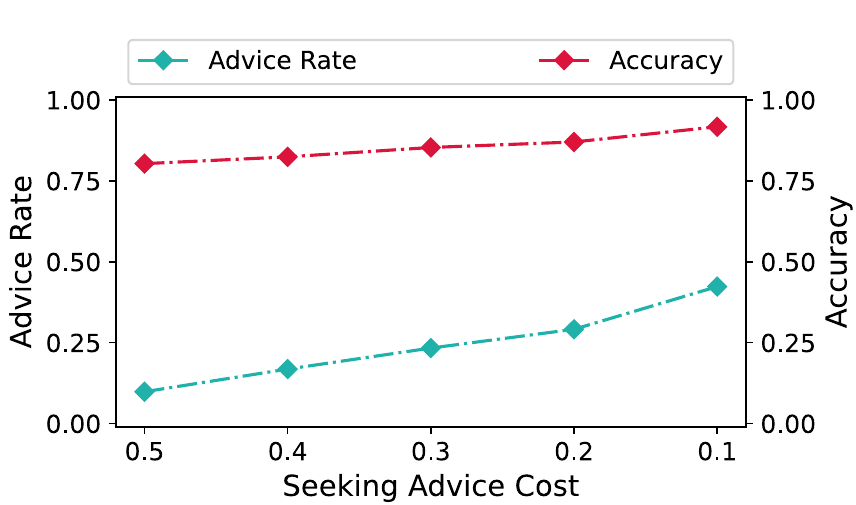}
\vspace{-0.5cm}
\caption{Advice rate, accuracy along with seeking advice cost $c$ on ProductQA.}
\label{seek_advice_cost_plot}
\end{minipage}
\end{figure}

As Table~\ref{overall_result} shows, our AGILE agent outperforms all baselines on ProductQA. Notably, the average total score of \texttt{agile-vic13b-ppo} across six test groups shows a relative improvement of 9.2\% in short answers and 5.0\% in long answers to \texttt{agile-gpt4-prompt} where the seeking advice cost is added into the prompt. Concretely, \texttt{agile-vic13b-ppo} uses a comparable number of seeking advice to achieve 7.4\% higher accuracy in short answers than \texttt{agile-gpt4-prompt}, and as Figure~\ref{fig:acc_sa_comparison} shows, this accuracy improvement is consistent across the whole trajectory. Our \texttt{agile-vic7b-ppo} agent also outperforms \texttt{agile-gpt4-prompt} in average total scores. Note that the GPT-4 agent knows the seeking advice cost from its prompt (see Figure~\ref{fig:agile_gpt_productqa_prompt}).

We investigate the impact of varying the seeking advice cost. As shown in Figure~\ref{seek_advice_cost_plot}, when the cost decreases, both the advice rate and the accuracy increase, indicating greater utilization of human assistance. Specifically, with a high cost of 0.5, the advice rate is close to 0, and at a low cost of 0.1, the accuracy is close to 1. This result demonstrates that by adjusting the cost and through RL training, we can effectively manage the trade-off between accuracy and human cost. For instance, the agent can achieve 94.1\% accuracy on the Motherboards task with a seeking advice cost of $c=0.1$ (refer to Table \ref{ablation_reward}). This capability is especially important in realistic scenarios that demand high accuracy levels. In most experiments, we set the cost at a medium level with $c=0.3$.

To validate the accuracy of GPT-4 evaluator in assessing the long answer results, we randomly select 100 triplets (questions, reference long answer, model-predicted long answer) and manually labeled the correctness. The results show a 94\% agreement rate between the GPT-4 evaluator and the author.

\paragraph{Ablation study} 
\begin{table}[htbp]
\renewcommand{\arraystretch}{1}
\centering
\caption{Ablation studies for disabling reflection, memory, seeking advice, tool use, or RL training. Here, \texttt{non-adapt-advice} means that seeking advice is invoked for the first $K$ sessions of the trajectory, where $K$ equals to the number of \texttt{[SeekAdvice]} performed by \texttt{agile-vic13b-ppo}. See Table~\ref{ablation_appendix} for ablation results on individual product groups.}
\label{ablation}
\scalebox{0.9}{
\begin{tabular}{lccc}
\toprule[1pt]
\multicolumn{1}{l}{\textbf{Method}} & \textbf{Advice Rate}~$\downarrow$ & \textbf{Accuracy}~$\uparrow$ & \textbf{Total Score}~$\uparrow$ \\
\midrule[0.5pt]
w/o Reflection & 0.270 & 0.852 & 0.771{\tiny{(-1.7\%)}} \\
w/o Memory & 0.407 & 0.876 & 0.754{\tiny{(-4.0\%)}} \\
w/o Advice & 0.000 & 0.747 & 0.747{\tiny{(-5.0\%)}} \\
non-adapt-advice & 0.233 & 0.812 & 0.742{\tiny{(-5.7\%)}} \\
w/o Tool-Use & 0.492 & 0.864 & 0.717{\tiny{(-9.3\%)}} \\
w/o RL & 0.256 & 0.843 & 0.766{\tiny{(-2.3\%)}} \\
\midrule[0.5pt]
agile-vic13b-ppo (ours) & 0.233 & \textbf{0.854} & \textbf{0.784} \\
\bottomrule[1pt]
\end{tabular}
}
\vspace{-0.2cm}
\end{table}

We present ablation studies in Table~\ref{ablation} to assess the contributions of individual agent components and the effects of RL training. The table indicates that disabling the option to seek advice (w/o Advice) leads to a 10.7\% drop in accuracy and a 5.0\% relative reduction in total score. Forcing the agent to seek advice at the initial part of the trajectory (Non-adapt Advice) causes a 4.2\% decrease in accuracy, underscoring the value of adaptive decision-making. Removing reflection and memory capabilities (w/o Memory and w/o Reflection) both increase the frequency of advice-seeking, as the agent struggles to accumulate or leverage valuable knowledge, consequently decreasing the total score. Furthermore, disabling tool use (w/o Tool-Use) causes a substantial 25.9\% increase in the advice-seeking rate because the agent's capabilities are diminished, making it more reliant on external advice. Lastly, RL training improves the relative total score by 2.3\%, lowers the advice-seeking rate, and boosts accuracy, demonstrating that RL training effectively optimizes the policy. Additional results on RL training can be found in Appendix~\ref{ppo_training}.

In Appendix~\ref{app:case_study}, we present detailed examples of \texttt{agile-vic13b-ppo} illustrating how memory, tools, seeking advice, and reflection enhance the agent workflow.

\begin{wrapfigure}{r}{0.43\textwidth}
\vspace{-0.5cm}
\centering
\includegraphics[width=0.41\textwidth]{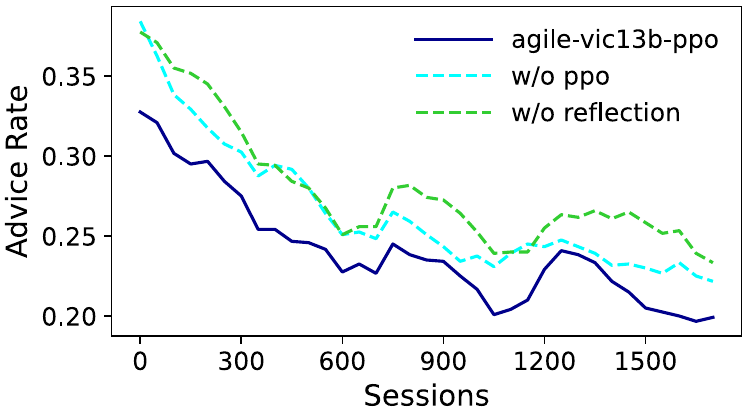}
\vspace{-0.3cm}
\caption{Advice rate over the following 200 sessions on ProductQA ($c = 0.3$).
}
\label{fig:seek_advice_rate}
\vspace{-0.5cm}
\end{wrapfigure}

\paragraph{Trend of advice rate} Figure~\ref{fig:seek_advice_rate} demonstrates a consistent decrease in the advice rate of \texttt{agile-vic13b-ppo} as more sessions are added to the trajectory. This decline can be attributed to the agent progressively accumulating knowledge and becoming more independent. Additionally, the figure illustrates that disabling RL training or reflection leads to a significant increase in the advice rate, underscoring the importance of RL training and reflection in reducing human costs.

\subsection{Results on MedMCQA}
\vspace{-10pt}
\begin{table}[htbp]
\renewcommand{\arraystretch}{1}
\centering
\caption{Results on the {MedMCQA} dev dataset. \texttt{X-prompt} represents directly prompting the model X; \texttt{agile-X-Y} represents incorporating the model X within the AGILE framework, while Y represents prompting, ablation studies or standard PPO training. The seeking advice cost is $c=0.4$.}
\label{overall_result_mcqa}
\scalebox{0.9}{
\begin{tabular}{lccc}
\toprule[1pt]
\multicolumn{1}{l}{\textbf{Method}} & \textbf{Advice Rate}~$\downarrow$ & \textbf{Accuracy}~$\uparrow$ & \textbf{Total Score}~$\uparrow$ \\
\midrule[0.5pt]
Meerkat-7b-prompt & - & 0.534 & - \\
gpt3.5-prompt\cite{nori2023capabilities} & - & 0.501 & - \\
gpt4-prompt\cite{nori2023capabilities} & - & 0.695 & - \\
gpt4-Medprompt\cite{nori2023generalist} & - & 0.791 & - \\
\midrule[0.5pt]
agile-gpt3.5-prompt & 0.194 & 0.697 & 0.619 \\
agile-gpt4-prompt & 0.421 & \textbf{0.884} & 0.721 \\
\midrule[0.5pt]
agile-mek7b-w/o Reflection & 0.368 & 0.790 & 0.643 \\
agile-mek7b-w/o Memory & 0.506 & 0.741 & 0.539 \\
agile-mek7b-w/o Advice & 0.000 & 0.620 & 0.620 \\
agile-mek7b-w/o RL & 0.322 & 0.837 & 0.708 \\
\midrule[0.5pt]
agile-mek7b-ppo (ours) & 0.316 & 0.852 & \textbf{0.726} \\
\bottomrule[1pt]
\end{tabular}
}
\vspace{-0.2cm}
\end{table}

Our \texttt{agile-mek7b-ppo} agent, based on the smaller Meerkat-7b \cite{kim2024small} model, reaches an accuracy of 85.2\% with an advice rate of 31.6\%. As Table \ref{overall_result_mcqa} shows, this represents a 31.8\% accuracy increase over the base model \texttt{Meerkat-7b-prompt} and a 6.1\% increase over the state-of-the-art \texttt{gpt4-Medprompt} \cite{nori2023generalist}. Table \ref{overall_result_mcqa} also shows that the ability to seek advice alone contributes a 23.2\% accuracy gain, meaning that each instance of seeking advice corrects an average of 0.73 prediction errors. This indicates that PPO training effectively helps the agent identify its mistakes. For a fair comparison, we also evaluate \texttt{agile-gpt3.5-prompt} and \texttt{agile-gpt4-prompt}, which incorporate GPT-3.5 and GPT-4 within our AGILE framework. These agents also leverage advice-seeking to enhance accuracy, but without RL training, their total scores are lower than \texttt{agile-mek7b-ppo}. Finally, through ablation studies, we confirmed the essential roles of memory, reflection, seeking advice, and RL training in achieving high performance. Removing these components leads to a significant drop in total scores, detailed in Table \ref{overall_result_mcqa}.

\subsection{Results on HotPotQA}

We compare our method against several baselines. Specifically, we found the original ReAct baseline implementation in \cite{yao2022react} to be suboptimal. By reproducing their results with GPT-4 (ReAct-gpt4-prompt), we observed improved performance. As shown in Table~\ref{overall_result_hotpot}, our agile agent outperforms all baselines in accuracy, achieving a 40.0\% relative improvement over ReAct-gpt4-prompt, which is the strongest baseline. Additionally, compared to \texttt{agile-gpt4-prompt}, the trained \texttt{agile-vic13b-ppo} demonstrates both higher accuracy and a lower advice rate, leading to a 10.8\% relative increase in total score. Ablation studies confirm that removing either seeking-advice or PPO training results in a significant decrease in the total score.

\begin{table}[htbp]
\vspace{-10pt}
\renewcommand{\arraystretch}{1}
\centering
\caption{Results on the HotPotQA full dev dataset. \texttt{X-prompt} represents directly prompting the model X; \texttt{agile-X-Y} represents incorporating the model X within the AGILE framework, while Y represents prompting, ablation studies or standard PPO training. The seeking advice cost is $c=0.3$.}
\label{overall_result_hotpot}
\scalebox{0.9}{
\begin{tabular}{lcccc}
\toprule[1pt]
\multicolumn{1}{l}{\textbf{Method}} & \textbf{Advice Rate}~$\downarrow$ & 
\begin{tabular}[c]{@{}c@{}}\textbf{Accuracy}~$\uparrow$ \\\textbf{(Exact Match)}\end{tabular}
& \begin{tabular}[c]{@{}c@{}}\textbf{Accuracy}~$\uparrow$ \\\textbf{(GPT-4 Evaluator)}\end{tabular}
& \begin{tabular}[c]{@{}c@{}}\textbf{Total Score}~$\uparrow$ \\\textbf{(Exact Match)}\end{tabular}
\\
\midrule[0.5pt]
ReAct~\cite{yao2022react} & - & 0.351 & - & - \\
ReAct-gpt4-prompt & - & 0.482 & - & - \\
CRITIC~\cite{goucritic} & - & 0.443 & - & - \\
Expel~\cite{zhao2024expel} & - & 0.390 & - & - \\
AutoAct~\cite{qiao2024autoact} & - & 0.384 & - & - \\
\midrule[0.5pt]
agile-gpt4-prompt & 0.194 & 0.664 & 0.842 & 0.567 \\
\midrule[0.5pt]
agile-vic13b-w/o Advice & 0.000 & 0.553 & 0.751 & 0.553 \\
agile-vic13b-w/o RL & 0.171 & 0.668 & 0.857 & 0.617 \\
\midrule[0.5pt]
agile-vic13b-ppo (ours) & 0.156 & {\bf 0.675} & {\bf 0.858} & {\bf 0.628} \\
\bottomrule[1pt]
\end{tabular}
}
\end{table}


\section{Related work}

\begin{table}[ht]
\vspace{-10pt}
\centering
\caption{Related work on LLM agents. \agentnm stands out as the pioneering work that trains the entire agent using reinforcement learning, incorporating proactive human advice-seeking.}
\label{tab:llm_agent_comparison}
\scalebox{0.8}{
\begin{tabular}{llcccccc}
\toprule[1pt]
\multicolumn{1}{l}{\textbf{LLM Agent}} &
\multicolumn{1}{l}{\textbf{LLM}} &
\textbf{SFT} &
\textbf{RL}&
\textbf{Memory} &
\textbf{Tools} &
\textbf{Reflection} &
\begin{tabular}[c]{@{}c@{}}\textbf{Proactive}\\ \textbf{Human-agent}\\ \textbf{Interaction}\end{tabular} \\
\midrule[0.5pt]
WebGPT~\cite{nakano2021webgpt}  & GPT-3 &\myCheckMark & \myCheckMark & \myCrossMark & \myCheckMark  &   \myCrossMark    & \myCrossMark
\\
ReAct~\cite{yao2022react}             & PaLM-540b          & \myCheckMark & \myCrossMark  & \myCheckMark & \myCheckMark & \myCrossMark  & \myCrossMark  \\

Reflexion~\cite{shinn2024reflexion}         & GPT-3/3.5/4        & \myCrossMark  & \myCrossMark & \myCheckMark & \myCheckMark & \myCheckMark & \myCrossMark  \\

ChatDev~\cite{qian2023communicative}           & ChatGPT-turbo-16k  & \myCrossMark  & \myCrossMark  & \myCheckMark & \myCheckMark & \myCheckMark & \myCrossMark  \\
RAP~\cite{kagaya2024rap}               & LLaMA-33b         & \myCrossMark  & \myCrossMark & \myCheckMark & \myCrossMark  & \myCrossMark  & \myCrossMark  \\
AutoAct~\cite{qiao2024autoact}           & LLaMA2-70b & \myCheckMark  & \myCrossMark & \myCheckMark  & \myCheckMark  & \myCheckMark & \myCrossMark\\
TPTU~\cite{ruan2023tptu}           & ChatGPT/InternLM & \myCrossMark  & \myCrossMark & \myCheckMark  & \myCheckMark  & \myCheckMark & \myCrossMark\\
\midrule[0.5pt]
\agentnm (Ours)              & Vicuna-13b/Meerkat-7b         & \myCheckMark & \myCheckMark & \myCheckMark & \myCheckMark & \myCheckMark & \myCheckMark \\
\bottomrule[1pt]
\end{tabular}
}
\vspace{-10pt}
\end{table}

\paragraph{LLM agents} Large Language Models (LLMs) have demonstrated substantial capabilities in following instructions, reasoning, and planning. Numerous research works, as shown in Table~\ref{tab:llm_agent_comparison}, utilizing prompt engineering, have constructed remarkable LLM agents capable of autonomously resolving complex tasks across various environments~\cite{park2023generative,wang2023describe,bran2023chemcrow,qian2023communicative,chen2023autoagents}. Furthermore, extensive works identify key components in the design of LLM agents, including planning~\cite{nakano2021webgpt,shen2024hugginggpt,hao2023reasoning,qiao2024autoact,yao2022react,ruan2023tptu}, tool-use~\cite{lewis2020retrieval,patil2023gorilla,yang2023mm,schick2024toolformer}, and reflection~\cite{shinn2024reflexion,madaan2024self}. In this work, we enable the agent to utilize memory, tools and proactively learn from the environment. We then formulate the entire process within an RL framework so that all agent skills can be jointly optimized end-to-end.

\paragraph{Human-agent interaction} Although LLMs face practical challenges, such as hallucination~\cite{zhang2023siren} and a lack of long-tail knowledge~\cite{kandpal2023large}, consulting human experts can help mitigate these issues. Several studies~\cite{zhang2023ask,xiao2023llm} have incorporated human experts into agent workflows relying on passive feedback or predefined rules. However, these approaches do not involve proactively seeking advice, which requires more complex decision-making. While \cite{chen2023asking,qian2024tell} train models to ask questions using behavior cloning, they ignore the fact that the decision to seek advice must be based on the LLM's own knowledge and capabilities~\cite{zhou2023navigating,kuhn2022semantic,kadavath2022language}. \cite{ren2023robots} use a calibrated version of an LLM's token probabilities as a confidence measure, yet token probabilities tend to be overconfident~\cite{xiong2023can}, and existing calibration methods don't generalize well to our agent setting when the LLM makes multiple decisions in sequence. Ultimately, the challenge of seeking advice is tied to the LLM's self-evaluation, which is difficult to ground truth or optimize through SFT. In our RL framework, the value and cost of seeking advice can be directly represented as RL rewards, enabling the proactive skill of seeking advice to be optimized as part of the policy model on end-to-end RL training.

\paragraph{LLM agent benchmarks} Several benchmarks have been designed to assess the capabilities of agents. For instance, the Webshop~\cite{yao2022webshop} and Mind2Web~\cite{deng2024mind2web} datasets evaluate agents' tool usage and planning abilities within a web environment. HotPotQA~\cite{yang2018hotpotqa} and TriviaQA~\cite{joshi2017triviaqa} focus on agents' reasoning and tool usage for question answering. ALFWorld~\cite{shridhar2020alfworld} examines planning and navigation skills, while ScienceWorld~\cite{wang-etal-2022-scienceworld} provides an interactive text-based environment to evaluate agents’ scientific aptitude. As illustrated in Table~\ref{tab:llm_agent_benchmarks}, despite these existing benchmarks, none comprehensively addresses all the core challenges of real-world agent applications, such as handling long-tail knowledge, human-agent interaction, long-term memory usage, tool usage, self-evaluation, and reflection. This motivated us to develop ProductQA.

\begin{table}[ht]
\vspace{-10pt}
\centering
\caption{Benchmarks for evaluating LLM agents. ProductQA features long trajectories, tool use, long-term knowledge accumulation, and cross-task capabilities.}
\label{tab:llm_agent_benchmarks}
\scalebox{0.88}{
\begin{tabular}{lccccccc}
\toprule[1pt]
\textbf{Datasets}
& \textbf{Type}
& \textbf{Fields}
& \textbf{Size}
& \begin{tabular}[c]{@{}c@{}}\textbf{Long} \\\textbf{Trajectory}\end{tabular}
& \begin{tabular}[c]{@{}c@{}}\textbf{Tool} \\\textbf{Usage}\end{tabular}
& \begin{tabular}[c]{@{}c@{}}\textbf{Long-term} \\\textbf{Knowledge}\end{tabular}
& \begin{tabular}[c]{@{}c@{}}\textbf{Cross} \\\textbf{Task}\end{tabular}\\
\midrule[0.5pt]
Webshop~\cite{yao2022webshop}     & Simulator & Web        & 12,087  & \myCrossMark           & \myCrossMark      & \myCrossMark               & \myCrossMark        \\
Mind2Web~\cite{deng2024mind2web}  & Simulator & Web        & 2,350   & \myCrossMark           & \myCrossMark      & \myCrossMark               & \myCheckMark       \\
ALFWorld~\cite{shridhar2020alfworld}            & Simulator & Navigation & 3,827  & \myCrossMark & \myCrossMark & \myCrossMark & \myCheckMark \\
ScienceWorld~\cite{wang-etal-2022-scienceworld} & Simulator & Science    & 7,207 & \myCrossMark & \myCrossMark & \myCrossMark & \myCrossMark  \\
HotPotQA~\cite{yang2018hotpotqa}  & QA        & Wikipedia  & 112,779 & \myCrossMark           & \myCheckMark     & \myCrossMark               & \myCrossMark        \\
TriviaQA~\cite{joshi2017triviaqa} & QA        & Web        & 95,956  & \myCrossMark           & \myCheckMark     & \myCheckMark              & \myCrossMark        \\
\midrule[0.5pt]
ProductQA (ours)                                        & QA        & E-commerce & 88,229    & \myCheckMark          & \myCheckMark     & \myCheckMark              & \myCheckMark   \\
\bottomrule[1pt]
\end{tabular}
}
\vspace{-10pt}
\end{table}

\section{Conclusion and future work}

In this work, we introduce a novel reinforcement learning framework of LLM agents, called \agentnm. First, the whole system of \agentnm is trained end-to-end by reinforcement learning. Second, \agentnm has the ability of seeking advice from external human experts. In addition, we develop a challenging dataset of complex QA, ProductQA, for comprehensive evaluation of an agent's capabilities. Extensive experiments demonstrate that within our framework, an agent based on a smaller model after RL training can outperform GPT-4.

\agentnm is a general agent framework and we can certainly consider multiple extensions of it. An agent can be equipped with more tools, such as multimodal perception, manipulations in physical environments, logical reasoning, among others.
We posit that \agentnm 's activities can be categorized into two distinct types: utilizing its LLM alone, and integrating the LLM with other tools. These two approaches conceptually align with the human cognitive processes known as System 1 and System 2~\cite{kahneman2003maps,bengio2021deep}. Furthermore, \agentnm 's memory serves as a repository for the accumulation of experiences and knowledge, which is crucial for self-improvement. Consequently, \agentnm offers an architecture for an very powerful agent that has the potential to attain human-level intelligence.

\agentnm also includes interactions between the agent and external human experts. The framework can be extended to allow interactions with humans or machine agents in various roles such as students or teachers, and in different formats such as debates or coordination. Furthermore, \agentnm can be employed in multi-agent systems.

\section*{Acknowledgements}
The authors thank anonymous reviewers for their helpful suggestions.

\bibliographystyle{plain}
\bibliography{main}

\begin{thebibliography}{10}

\bibitem{bengio2021deep}
Yoshua Bengio, Yann Lecun, and Geoffrey Hinton.
\newblock Deep learning for ai.
\newblock {\em Communications of the ACM}, 64(7):58--65, 2021.

\bibitem{bran2023chemcrow}
Andres~M Bran, Sam Cox, Oliver Schilter, Carlo Baldassari, Andrew~D White, and Philippe Schwaller.
\newblock Chemcrow: Augmenting large-language models with chemistry tools.
\newblock {\em arXiv preprint arXiv:2304.05376}, 2023.

\bibitem{brown2020language}
Tom Brown, Benjamin Mann, Nick Ryder, Melanie Subbiah, Jared~D Kaplan, Prafulla Dhariwal, Arvind Neelakantan, Pranav Shyam, Girish Sastry, Amanda Askell, et~al.
\newblock Language models are few-shot learners.
\newblock {\em Advances in neural information processing systems}, 33:1877--1901, 2020.

\bibitem{chen2023autoagents}
Guangyao Chen, Siwei Dong, Yu~Shu, Ge~Zhang, Jaward Sesay, B{\"o}rje~F Karlsson, Jie Fu, and Yemin Shi.
\newblock Autoagents: A framework for automatic agent generation.
\newblock {\em arXiv preprint arXiv:2309.17288}, 2023.

\bibitem{chen2023asking}
Xiaoyu Chen, Shenao Zhang, Pushi Zhang, Li~Zhao, and Jianyu Chen.
\newblock Asking before action: Gather information in embodied decision making with language models.
\newblock {\em arXiv preprint arXiv:2305.15695}, 2023.

\bibitem{vicuna2023}
Wei-Lin Chiang, Zhuohan Li, Zi~Lin, Ying Sheng, Zhanghao Wu, Hao Zhang, Lianmin Zheng, Siyuan Zhuang, Yonghao Zhuang, Joseph~E. Gonzalez, Ion Stoica, and Eric~P. Xing.
\newblock Vicuna: An open-source chatbot impressing gpt-4 with 90\%* chatgpt quality, March 2023.

\bibitem{deng2024mind2web}
Xiang Deng, Yu~Gu, Boyuan Zheng, Shijie Chen, Sam Stevens, Boshi Wang, Huan Sun, and Yu~Su.
\newblock Mind2web: Towards a generalist agent for the web.
\newblock {\em Advances in Neural Information Processing Systems}, 36, 2024.

\bibitem{deng2023product}
Yang Deng, Wenxuan Zhang, Qian Yu, and Wai Lam.
\newblock Product question answering in e-commerce: A survey.
\newblock In {\em Proceedings of the 61st Annual Meeting of the Association for Computational Linguistics (Volume 1: Long Papers)}, pages 11951--11964, 2023.

\bibitem{goucritic}
Zhibin Gou, Zhihong Shao, Yeyun Gong, Yujiu Yang, Nan Duan, Weizhu Chen, et~al.
\newblock Critic: Large language models can self-correct with tool-interactive critiquing.
\newblock In {\em The Twelfth International Conference on Learning Representations}, 2024.

\bibitem{hao2023reasoning}
Shibo Hao, Yi~Gu, Haodi Ma, Joshua~Jiahua Hong, Zhen Wang, Daisy~Zhe Wang, and Zhiting Hu.
\newblock Reasoning with language model is planning with world model.
\newblock {\em arXiv preprint arXiv:2305.14992}, 2023.

\bibitem{jin2021disease}
Di~Jin, Eileen Pan, Nassim Oufattole, Wei-Hung Weng, Hanyi Fang, and Peter Szolovits.
\newblock What disease does this patient have? a large-scale open domain question answering dataset from medical exams.
\newblock {\em Applied Sciences}, 11(14):6421, 2021.

\bibitem{joshi2017triviaqa}
Mandar Joshi, Eunsol Choi, Daniel~S Weld, and Luke Zettlemoyer.
\newblock Triviaqa: A large scale distantly supervised challenge dataset for reading comprehension.
\newblock {\em arXiv preprint arXiv:1705.03551}, 2017.

\bibitem{kadavath2022language}
Saurav Kadavath, Tom Conerly, Amanda Askell, Tom Henighan, Dawn Drain, Ethan Perez, Nicholas Schiefer, Zac Hatfield-Dodds, Nova DasSarma, Eli Tran-Johnson, et~al.
\newblock Language models (mostly) know what they know.
\newblock {\em arXiv preprint arXiv:2207.05221}, 2022.

\bibitem{kagaya2024rap}
Tomoyuki Kagaya, Thong~Jing Yuan, Yuxuan Lou, Jayashree Karlekar, Sugiri Pranata, Akira Kinose, Koki Oguri, Felix Wick, and Yang You.
\newblock Rap: Retrieval-augmented planning with contextual memory for multimodal llm agents.
\newblock {\em arXiv preprint arXiv:2402.03610}, 2024.

\bibitem{kahneman2003maps}
Daniel Kahneman.
\newblock Maps of bounded rationality: Psychology for behavioral economics.
\newblock {\em American economic review}, 93(5):1449--1475, 2003.

\bibitem{kandpal2023large}
Nikhil Kandpal, Haikang Deng, Adam Roberts, Eric Wallace, and Colin Raffel.
\newblock Large language models struggle to learn long-tail knowledge.
\newblock In {\em International Conference on Machine Learning}, pages 15696--15707. PMLR, 2023.

\bibitem{kim2024small}
Hyunjae Kim, Hyeon Hwang, Jiwoo Lee, Sihyeon Park, Dain Kim, Taewhoo Lee, Chanwoong Yoon, Jiwoong Sohn, Donghee Choi, and Jaewoo Kang.
\newblock Small language models learn enhanced reasoning skills from medical textbooks.
\newblock {\em arXiv preprint arXiv:2404.00376}, 2024.

\bibitem{kuhn2022semantic}
Lorenz Kuhn, Yarin Gal, and Sebastian Farquhar.
\newblock Semantic uncertainty: Linguistic invariances for uncertainty estimation in natural language generation.
\newblock In {\em The Eleventh International Conference on Learning Representations}, 2022.

\bibitem{lewis2020retrieval}
Patrick Lewis, Ethan Perez, Aleksandra Piktus, Fabio Petroni, Vladimir Karpukhin, Naman Goyal, Heinrich K{\"u}ttler, Mike Lewis, Wen-tau Yih, Tim Rockt{\"a}schel, et~al.
\newblock Retrieval-augmented generation for knowledge-intensive nlp tasks.
\newblock {\em Advances in Neural Information Processing Systems}, 33:9459--9474, 2020.

\bibitem{liang2022holistic}
Percy Liang, Rishi Bommasani, Tony Lee, Dimitris Tsipras, Dilara Soylu, Michihiro Yasunaga, Yian Zhang, Deepak Narayanan, Yuhuai Wu, Ananya Kumar, et~al.
\newblock Holistic evaluation of language models.
\newblock {\em arXiv preprint arXiv:2211.09110}, 2022.

\bibitem{madaan2024self}
Aman Madaan, Niket Tandon, Prakhar Gupta, Skyler Hallinan, Luyu Gao, Sarah Wiegreffe, Uri Alon, Nouha Dziri, Shrimai Prabhumoye, Yiming Yang, et~al.
\newblock Self-refine: Iterative refinement with self-feedback.
\newblock {\em Advances in Neural Information Processing Systems}, 36, 2024.

\bibitem{nakano2021webgpt}
Reiichiro Nakano, Jacob Hilton, Suchir Balaji, Jeff Wu, Long Ouyang, Christina Kim, Christopher Hesse, Shantanu Jain, Vineet Kosaraju, William Saunders, et~al.
\newblock Webgpt: Browser-assisted question-answering with human feedback.
\newblock {\em arXiv preprint arXiv:2112.09332}, 2021.

\bibitem{ni2019justifying}
Jianmo Ni, Jiacheng Li, and Julian McAuley.
\newblock Justifying recommendations using distantly-labeled reviews and fine-grained aspects.
\newblock In {\em Proceedings of the 2019 conference on empirical methods in natural language processing and the 9th international joint conference on natural language processing (EMNLP-IJCNLP)}, pages 188--197, 2019.

\bibitem{nori2023capabilities}
Harsha Nori, Nicholas King, Scott~Mayer McKinney, Dean Carignan, and Eric Horvitz.
\newblock Capabilities of gpt-4 on medical challenge problems.
\newblock {\em arXiv preprint arXiv:2303.13375}, 2023.

\bibitem{nori2023generalist}
Harsha Nori, Yin~Tat Lee, Sheng Zhang, Dean Carignan, Richard Edgar, Nicolo Fusi, Nicholas King, Jonathan Larson, Yuanzhi Li, Weishung Liu, Renqian Luo, Scott~Mayer McKinney, Robert~Osazuwa Ness, Hoifung Poon, Tao Qin, Naoto Usuyama, Chris White, and Eric Horvitz.
\newblock Can generalist foundation models outcompete special-purpose tuning? case study in medicine, 2023.

\bibitem{openai2023gpt4}
OpenAI.
\newblock Gpt-4 technical report.
\newblock {\em arXiv preprint arXiv:2303.08774}, 2023.

\bibitem{pal2022medmcqa}
Ankit Pal, Logesh~Kumar Umapathi, and Malaikannan Sankarasubbu.
\newblock Medmcqa: A large-scale multi-subject multi-choice dataset for medical domain question answering.
\newblock In {\em Conference on health, inference, and learning}, pages 248--260. PMLR, 2022.

\bibitem{park2023generative}
Joon~Sung Park, Joseph O'Brien, Carrie~Jun Cai, Meredith~Ringel Morris, Percy Liang, and Michael~S Bernstein.
\newblock Generative agents: Interactive simulacra of human behavior.
\newblock In {\em Proceedings of the 36th Annual ACM Symposium on User Interface Software and Technology}, pages 1--22, 2023.

\bibitem{patil2023gorilla}
Shishir~G Patil, Tianjun Zhang, Xin Wang, and Joseph~E Gonzalez.
\newblock Gorilla: Large language model connected with massive apis.
\newblock {\em arXiv preprint arXiv:2305.15334}, 2023.

\bibitem{qian2023communicative}
Chen Qian, Xin Cong, Cheng Yang, Weize Chen, Yusheng Su, Juyuan Xu, Zhiyuan Liu, and Maosong Sun.
\newblock Communicative agents for software development.
\newblock {\em arXiv preprint arXiv:2307.07924}, 2023.

\bibitem{qian2024tell}
Cheng Qian, Bingxiang He, Zhong Zhuang, Jia Deng, Yujia Qin, Xin Cong, Yankai Lin, Zhong Zhang, Zhiyuan Liu, and Maosong Sun.
\newblock Tell me more! towards implicit user intention understanding of language model driven agents.
\newblock {\em arXiv preprint arXiv:2402.09205}, 2024.

\bibitem{qiao2024autoact}
Shuofei Qiao, Ningyu Zhang, Runnan Fang, Yujie Luo, Wangchunshu Zhou, Yuchen~Eleanor Jiang, Chengfei Lv, and Huajun Chen.
\newblock Autoact: Automatic agent learning from scratch via self-planning.
\newblock {\em arXiv preprint arXiv:2401.05268}, 2024.

\bibitem{reimers2019sentencebert}
Nils Reimers and Iryna Gurevych.
\newblock Sentence-bert: Sentence embeddings using siamese bert-networks, 2019.

\bibitem{ren2023robots}
Allen~Z Ren, Anushri Dixit, Alexandra Bodrova, Sumeet Singh, Stephen Tu, Noah Brown, Peng Xu, Leila Takayama, Fei Xia, Jake Varley, et~al.
\newblock Robots that ask for help: Uncertainty alignment for large language model planners.
\newblock {\em arXiv preprint arXiv:2307.01928}, 2023.

\bibitem{ruan2023tptu}
Jingqing Ruan, Yihong Chen, Bin Zhang, Zhiwei Xu, Tianpeng Bao, Guoqing Du, Shiwei Shi, Hangyu Mao, Xingyu Zeng, and Rui Zhao.
\newblock Tptu: Task planning and tool usage of large language model-based ai agents.
\newblock {\em arXiv preprint arXiv:2308.03427}, 2023.

\bibitem{schick2024toolformer}
Timo Schick, Jane Dwivedi-Yu, Roberto Dess{\`\i}, Roberta Raileanu, Maria Lomeli, Eric Hambro, Luke Zettlemoyer, Nicola Cancedda, and Thomas Scialom.
\newblock Toolformer: Language models can teach themselves to use tools.
\newblock {\em Advances in Neural Information Processing Systems}, 36, 2024.

\bibitem{schulman2017proximal}
John Schulman, Filip Wolski, Prafulla Dhariwal, Alec Radford, and Oleg Klimov.
\newblock Proximal policy optimization algorithms.
\newblock {\em arXiv preprint arXiv:1707.06347}, 2017.

\bibitem{shen2023xpqa}
Xiaoyu Shen, Akari Asai, Bill Byrne, and Adri{\`a} de~Gispert.
\newblock xpqa: Cross-lingual product question answering across 12 languages.
\newblock {\em arXiv preprint arXiv:2305.09249}, 2023.

\bibitem{shen2024hugginggpt}
Yongliang Shen, Kaitao Song, Xu~Tan, Dongsheng Li, Weiming Lu, and Yueting Zhuang.
\newblock Hugginggpt: Solving ai tasks with chatgpt and its friends in hugging face.
\newblock {\em Advances in Neural Information Processing Systems}, 36, 2024.

\bibitem{shinn2024reflexion}
Noah Shinn, Federico Cassano, Ashwin Gopinath, Karthik Narasimhan, and Shunyu Yao.
\newblock Reflexion: Language agents with verbal reinforcement learning.
\newblock {\em Advances in Neural Information Processing Systems}, 36, 2024.

\bibitem{shridhar2020alfworld}
Mohit Shridhar, Xingdi Yuan, Marc-Alexandre Cote, Yonatan Bisk, Adam Trischler, and Matthew Hausknecht.
\newblock Alfworld: Aligning text and embodied environments for interactive learning.
\newblock In {\em International Conference on Learning Representations}, 2020.

\bibitem{wang2023voyager}
Guanzhi Wang, Yuqi Xie, Yunfan Jiang, Ajay Mandlekar, Chaowei Xiao, Yuke Zhu, Linxi Fan, and Anima Anandkumar.
\newblock Voyager: An open-ended embodied agent with large language models.
\newblock {\em arXiv preprint arXiv:2305.16291}, 2023.

\bibitem{wang-etal-2022-scienceworld}
Ruoyao Wang, Peter Jansen, Marc-Alexandre C{\^o}t{\'e}, and Prithviraj Ammanabrolu.
\newblock {S}cience{W}orld: Is your agent smarter than a 5th grader?
\newblock In Yoav Goldberg, Zornitsa Kozareva, and Yue Zhang, editors, {\em Proceedings of the 2022 Conference on Empirical Methods in Natural Language Processing}, pages 11279--11298, Abu Dhabi, United Arab Emirates, December 2022. Association for Computational Linguistics.

\bibitem{wang2023describe}
Zihao Wang, Shaofei Cai, Guanzhou Chen, Anji Liu, Xiaojian Ma, and Yitao Liang.
\newblock Describe, explain, plan and select: Interactive planning with large language models enables open-world multi-task agents.
\newblock {\em arXiv preprint arXiv:2302.01560}, 2023.

\bibitem{wei2022chain}
Jason Wei, Xuezhi Wang, Dale Schuurmans, Maarten Bosma, Fei Xia, Ed~Chi, Quoc~V Le, Denny Zhou, et~al.
\newblock Chain-of-thought prompting elicits reasoning in large language models.
\newblock {\em Advances in neural information processing systems}, 35:24824--24837, 2022.

\bibitem{xiao2023llm}
Hengjia Xiao and Peng Wang.
\newblock Llm a*: Human in the loop large language models enabled a* search for robotics.
\newblock {\em arXiv preprint arXiv:2312.01797}, 2023.

\bibitem{xiong2023can}
Miao Xiong, Zhiyuan Hu, Xinyang Lu, Yifei Li, Jie Fu, Junxian He, and Bryan Hooi.
\newblock Can llms express their uncertainty? an empirical evaluation of confidence elicitation in llms.
\newblock {\em arXiv preprint arXiv:2306.13063}, 2023.

\bibitem{yang2023mm}
Zhengyuan Yang, Linjie Li, Jianfeng Wang, Kevin Lin, Ehsan Azarnasab, Faisal Ahmed, Zicheng Liu, Ce~Liu, Michael Zeng, and Lijuan Wang.
\newblock Mm-react: Prompting chatgpt for multimodal reasoning and action.
\newblock {\em arXiv preprint arXiv:2303.11381}, 2023.

\bibitem{yang2018hotpotqa}
Zhilin Yang, Peng Qi, Saizheng Zhang, Yoshua Bengio, William~W Cohen, Ruslan Salakhutdinov, and Christopher~D Manning.
\newblock Hotpotqa: A dataset for diverse, explainable multi-hop question answering.
\newblock {\em arXiv preprint arXiv:1809.09600}, 2018.

\bibitem{yao2022webshop}
Shunyu Yao, Howard Chen, John Yang, and Karthik Narasimhan.
\newblock Webshop: Towards scalable real-world web interaction with grounded language agents.
\newblock {\em Advances in Neural Information Processing Systems}, 35:20744--20757, 2022.

\bibitem{yao2022react}
Shunyu Yao, Jeffrey Zhao, Dian Yu, Nan Du, Izhak Shafran, Karthik Narasimhan, and Yuan Cao.
\newblock React: Synergizing reasoning and acting in language models.
\newblock {\em arXiv preprint arXiv:2210.03629}, 2022.

\bibitem{zhang2023ask}
Qiang Zhang, Jason Naradowsky, and Yusuke Miyao.
\newblock Ask an expert: Leveraging language models to improve strategic reasoning in goal-oriented dialogue models.
\newblock {\em arXiv preprint arXiv:2305.17878}, 2023.

\bibitem{zhang2023siren}
Yue Zhang, Yafu Li, Leyang Cui, Deng Cai, Lemao Liu, Tingchen Fu, Xinting Huang, Enbo Zhao, Yu~Zhang, Yulong Chen, et~al.
\newblock Siren's song in the ai ocean: a survey on hallucination in large language models.
\newblock {\em arXiv preprint arXiv:2309.01219}, 2023.

\bibitem{zhao2024expel}
Andrew Zhao, Daniel Huang, Quentin Xu, Matthieu Lin, Yong-Jin Liu, and Gao Huang.
\newblock Expel: Llm agents are experiential learners.
\newblock In {\em Proceedings of the AAAI Conference on Artificial Intelligence}, volume~38, pages 19632--19642, 2024.

\bibitem{zhou2023navigating}
Kaitlyn Zhou, Dan Jurafsky, and Tatsunori Hashimoto.
\newblock Navigating the grey area: Expressions of overconfidence and uncertainty in language models.
\newblock {\em arXiv e-prints}, pages arXiv--2302, 2023.

\end{thebibliography}

\newpage
\appendix

\begin{center}
    \noindent\rule{\textwidth}{3pt} \vspace{-0.2cm}
    \LARGE \textbf{Appendix} 
    \noindent\rule{\textwidth}{1.2pt}
\end{center}

\startcontents[sections]
\printcontents[sections]{l}{1}{\setcounter{tocdepth}{2}}
\vspace{1.cm}
\newpage

\section{Session-level optimization algorithm}\label{algorithm}

Assume that the entire trajectory $\tau$ can be partitioned into sub-trajectories $(\tau_1, \tau_2, \cdots, \tau_n)$, each referred to as a \emph{session}. For session $i$, let $\mathcal{S}_i$ denote its initial state, where $c_i$ is the LLM context before the session starts, and $m_i$ is the memory before the session starts. In this section, we will explain how to transform a trajectory-level RL optimization algorithm into a session-level RL optimization algorithm.

Let $r(\tau)$ represent the total reward of trajectory $\tau$, and let $\pi_\theta$ be a policy parameterized by $\theta$. The optimization objective is to maximize the following expectation:
\begin{equation}\label{obj-0}
    R(\theta)=\mathbb{E}_{\tau\sim\pi_\theta}[r(\tau)].
\end{equation}

For an arbitrary session index $i$, the trajectory $\tau\sim\pi_\theta$ can be sampled in three stages: $\tau_{1:i-1}$, $\tau_i$, and $\tau_{i+1:n}$. These stages represent the sub-trajectory from session $1$ to $i-1$, the sub-trajectory for session $i$, and the sub-trajectory from session $i+1$ to $n$, respectively. Accordingly, we have
\begin{eqnarray}\label{obj-1}
    R(\theta)&=&\mathbb{E}_{\tau_{1:i-1}\sim\pi_\theta}\left[\mathbb{E}_{\tau_i\sim\pi_\theta(\cdot|\mathcal{S}_i)}\left[\mathbb{E}_{\tau_{i+1:n}\sim\pi_\theta(\cdot|\mathcal{S}_{i+1})}[r(\tau_{1:i-1})+r(\tau_i)+r(\tau_{i+1:n})]\right]\right]\nonumber\\
    &=&\mathbb{E}_{\tau_{1:i-1}\sim\pi_\theta}\left[r(\tau_{1:i-1}) + \mathbb{E}_{\tau_i\sim\pi_\theta(\cdot|\mathcal{S}_i)}\left[r(\tau_i) + V_{\pi_\theta}\left(\mathcal{S}_{i+1}\right)\right]\right].
\end{eqnarray}
Here, $\mathcal{S}_i$ and $\mathcal{S}_{i+1}$ denote the initial states of sessions $i$ and $i+1$ respectively. The term $r(\tau_{1:i-1})$ represents the total reward accumulated from session $1$ to $i-1$, while $r(\tau_i)$ is the reward obtained in session $i$. Additionally, $V_{\pi_\theta}\left(\mathcal{S}_{i+1}\right)$ represents the value function at state $\mathcal{S}_{i+1}$ with respect to policy $\pi_\theta$, indicating the expected total reward the agent expects to receive in the future. Averaging over all session indices, Eq.~\eqref{obj-1} gives:
\begin{equation}\label{obj-1-avg}
    R(\theta) = \frac{1}{n}\sum_{i=1}^n \mathbb{E}_{\tau_{1:i-1}\sim\pi_\theta}\left[r(\tau_{1:i-1}) + \mathbb{E}_{\tau_i\sim\pi_\theta(\cdot|\mathcal{S}_i)}\left[r(\tau_i) + V_{\pi_\theta}\left(\mathcal{S}_{i+1}\right)\right]\right].
\end{equation}

In Eq.~\eqref{obj-1-avg}, the parameter $\theta$ appears in three places -- two expectations and a value function -- making optimization challenging. To simplify the problem, we assume a base policy $\theta_k$ and define a proximal objective $R(\theta|\theta_k)$, where $\theta$ only appears in the session-level expectation:
\begin{equation}\label{obj-2}
    R(\theta|\theta_k)=\frac{1}{n}\sum_{i=1}^n\mathbb{E}_{\tau_{1:i-1}\sim\pi_\theta}\left[r(\tau_{1:i-1}) + \mathbb{E}_{\tau_i\sim\pi_\theta(\cdot|\mathcal{S}_i)}\left[r(\tau_i) + V_{\pi_{\theta_k}}\left(\mathcal{S}_{i+1}\right)\right]\right].
\end{equation}
$R(\theta|\theta_k)$ is an approximation to $R(\theta)$ in the neighborhood of $\theta_k$. If we employ an iterative optimization procedure:
\begin{enumerate}
\item Initialize $\theta_0$ from a reference policy (obtained through SFT).
\item For $k=0, 1, 2, \cdots$, compute $\theta_{k+1}\leftarrow \arg\max_\theta R(\theta|\theta_k)$.
\end{enumerate}
Then $\theta$ will converge to an (at least locally) optimal policy.

Now we are ready to illustrate why the optimization of $R(\theta|\theta_k)$ can be solved at the session level. Notice that
\begin{align*}\label{obj-3}
    R(\theta|\theta_k)&=\frac{1}{n}\sum_{i=1}^n\mathbb{E}_{\tau_{1:i-1}\sim\pi_{\theta_k}}\left[\mathbb{E}_{\tau_i\sim\pi_\theta(\cdot|\mathcal{S}_i)}[r(\tau_i)+V_{\pi_{\theta_k}}(\mathcal{S}_{i+1})-V_{\pi_{\theta_k}}(\mathcal{S}_i)]\right]\nonumber\\
    &\quad\quad\quad\quad + \mathbb{E}_{\tau_{1:i-1}\sim\pi_{\theta_k}}[r(\tau_{1:i-1}) + V_{\pi_{\theta_k}}(\mathcal{S}_i)]\\
    &=\frac{1}{n}\sum_{i=1}^n\mathbb{E}_{\tau_{1:i-1}\sim\pi_{\theta_k}}\left[\mathbb{E}_{\tau_i\sim\pi_\theta(\cdot|\mathcal{S}_i)}[r(\tau_i)+V_{\pi_{\theta_k}}(\mathcal{S}_{i+1})-V_{\pi_{\theta_k}}(\mathcal{S}_i)]\right] + \mathbb{E}_{\tau_i \sim\pi_{\theta_k}}[r(\tau)]
\end{align*}
On the right-hand side, the first term involves two sampling steps. The first step samples $\tau_{1:i-1}\sim\pi_{\theta_k}$. The inner terms inside the expectation only depends on $\mathcal{S}_i$, thus we can replace it by $\mathcal{S}_i\sim \pi_{\theta_k}$. The second term on the right-hand side is a constant independent of $\theta$. As a result, if we define a \emph{proxy reward}:
\begin{equation}\label{reward-0}
    \tilde{r}_k(\tau_i) := r(\tau_i) + (V_{\pi_{\theta_k}}(\mathcal{S}_{i+1}) - V_{\pi_{\theta_k}}(\mathcal{S}_i)).
\end{equation}
Then, we have
\begin{equation}\label{obj-4}
    R(\theta|\theta_k)=\frac{1}{n}\sum_{i=1}^n\mathbb{E}_{\mathcal{S}_i\sim\pi_{\theta_k}}\left[\mathbb{E}_{\tau_i\sim\pi_\theta(\cdot|\mathcal{S}_i)}[\tilde{r}_k(\tau_i)]\right] + \text{constant}.
\end{equation}
By Eq.~\eqref{obj-4}, $R(\theta|\theta_k)$ can be optimized by maximizing the average expected proxy reward for each session. The term $A_i := V_{\pi_{\theta_k}}(\mathcal{S}_{i+1}) - V_{\pi_{\theta_k}}(\mathcal{S}_i)$ measures the advantage of state $\mathcal{S}_{i+1}$ over state $\mathcal{S}_i$ with respect to a policy; thus, we call it the \emph{state advantage function}. This function can be either defined by heuristics or fitted by a neural network. In the latter case, one needs to sample trajectories from $\pi_{\theta_k}$, evaluate their rewards, and then use the (state, reward-to-go) pairs to train an estimator for the value function $V_{\pi_{\theta_k}}$.

\begin{algorithm}[t]
    \caption{Session-level optimization}
    \label{alg:session-level-optimization}
    \begin{algorithmic}[1]
    \State Initialize $\theta_0$ from a reference policy (obtained through SFT).
    \For{$k\gets 0, 1, 2, \cdots$}
        \State Sample a set of trajectories from $\pi_{\theta_k}$, denote the set by $T$.
        \State Define or fit a state advantage function from $T$.
        \For{each $\tau\in T$ }
            \State Partition it into sessions $(\tau_1, \tau_2, \cdots, \tau_n)$.
            \For{each $\tau_i$ }
            \State Evaluate $\tilde{r}_k(\tau_i)$ by Eq.~\eqref{reward-0} with the above state advantage function.
            \EndFor
        \EndFor
        \State Treat all sessions as independent, then employ an optimization algorithm (such as PPO) to obtain a new policy $\theta_{k+1}$ by maximizing Eq.~\eqref{obj-4}.
    \EndFor
    \end{algorithmic}
\end{algorithm}

Finally, we present the session-level optimization algorithm as Algorithm~\ref{alg:session-level-optimization}. In this algorithm, the state advantage function is the only component that concerns inter-session correlation. While the algorithm is iterative, we anticipate that in practice, the outer loop will require only a few iterations to converge.

\section{Implementation details of AGILE}\label{sec:product_qa_implementation_details}
\subsection{ProductQA}\label{sft_data_generation}

\paragraph{Implementation of \texttt{[GetQuestion]}} This function prompts the user for a new question and appends it to the LLM context. Every question is raised for a specific product, thus it has an associated product ID. Based on this ID, the function also appends the product information table's schema and the product metadata to the context.

\paragraph{Implementation of \texttt{[RetrieveMemory]}} This function employs the provided question as a query to retrieve the most relevant historical QA pair and the most relevant knowledge entry from the agent's memory. To safeguard sensitive data from sellers, the agent is restricted to accessing QA records exclusively for the queried product from historical interactions. However, it is permitted to retrieve general knowledge from the whole trajectory since this information is not seller-specific. We utilize an embedding-based retrieval method, specifically employing the all-MiniLM-L6-v2 model~\cite{reimers2019sentencebert} as the embedding model.

\paragraph{Implementation of \texttt{[SearchProdcut]}} This function utilizes the LLM to predict a SQL query based on the context, and then invoke a MySQL execution engine. It appends the result to the LLM context. If there is an execution error, then the error is appended to the context too.

\paragraph{Implementation of \texttt{[SeekAdvice]}} This requests for human expert advice and append it to the LLM context. In our implementation, the human expert simply returns the ground truth long answer from the ProductQA dataset.

\paragraph{Implementation of \texttt{[PredictAnswer]}} This function passes control to the LLM to continue generating a long answer and a short answer.

\paragraph{Implementation of \texttt{[Reflection]}} This function passes control to the LLM to continue generating a reflection result.

\paragraph{Training Data Generation} We generate training data on a session-by-session basis, where each session consists of a QA pair. A session begins with an initial memory, consisting of historical QA pairs and knowledge entries accumulated from previous sessions. Recall that the \texttt{[RetrieveMemory]} function retrieves only the most relevant QA pair and knowledge entry per session. Thus, in constructing training memories, it suffices to put the retrieved QA pair and the retrieved knowledge entry into the memory. We select them in the following stochastic way: the retrieved QA pair can be the most relevant QA pair from the training set, or a random QA pair, or omitted entirely; similarly for the retrieved knowledge entry.

Based on the initial memory, we generate trajectories by following the agent workflow detailed in Section~\ref{sec:exp-setting}. Each trajectory begins with \texttt{[GetUserQuestion]} and \texttt{[RetrieveMemory]}. For QAs classified as Search-QA, a \texttt{[SearchProduct]} function is appended, followed by the corresponding SQL query and its execution result. For other QA types, if an associated knowledge entry exists and is successfully retrieved, the trajectory will extend with a \texttt{[PredictAnswer]} call with the ground truth answer as its result. If the knowledge entry is not retrieved or is absent, we use GPT-4 to evaluate whether the question can be answered with the available context. If affirmative, a \texttt{[PredictAnswer]} with the ground truth answer is appended. Otherwise, the trajectory extends with a \texttt{[SeekAdvice]} call with the ground truth answer as the advice, and a \texttt{[Reflection]} call, where the reflection result is the knowledge entry if it exists, or "no information" if not. Then the reflection result is appended to the memory via \texttt{[UpdateMemory]}. Finally, the trajectory is concluded by \texttt{[SubmitAnswer]}.

In this way, we constructed 55,772 session-level trajectories in total, from 6 training tasks in ProductQA. This data is used for imitation learning. In PPO training, we reuse the initial memory data, while the session-level trajectories are generated by the model itself.

\subsection{MedMCQA}

For MedMCQA, the memory is initialized with all QA pairs from the training set, simulating that the agent has processed the training set before reaching the test set. We also add a knowledge entry for each QA pair, obtained through GPT-4 reflection (see Figure~\ref{fig:reflect_medmcqa_prompt} for the prompt).

\paragraph{Training data generation} We sample a subset of training data from MedMCQA to construct session-level trajectories. Each trajectory begins with \texttt{[GetUserQuestion]} and \texttt{[RetrieveMemory]}. The \texttt{[RetrieveMemory]} function retrieves the five most relevant QA pairs and pieces of knowledge from the initial memory, using the same embedding similarity search method employed in ProductQA. Then, we prompt GPT-4 to predict an answer with chain-of-thought reasoning. If the GPT-4 answer is correct, we append a \texttt{[PredictAnswer]} call, the GPT-4 chain-of-thought, and the ground-truth answer to the trajectory. If the GPT-4 answer is wrong, which suggests that the question is hard, we append a \texttt{[SeekAdvice]} call with the ground-truth answer, followed by a \texttt{[Reflection]} call with the reflection result generated by GPT-4. Then the reflection result is appended to the memory via \texttt{[UpdateMemory]}. Finally, the trajectory is concluded by \texttt{[SubmitAnswer]}. In this way, we obtain 23,015 session-level trajectories in total.

\subsection{HotPotQA}

In the HotPotQA task, the agent has the option to select either \texttt{[Search]}, \texttt{[SeekAdvice]} or \texttt{[PredictAnswer]} in each round. Following ReAct~\cite{yao2022react}, the agent first generates reasoning first and then selects an action.

\paragraph{Implementation of \texttt{[Search]}} This function uses the LLM to generate a search query and invokes a search API. The first result not already present in the LLM context is selected and appended to the existing context.

\paragraph{Training data generation} We use the HotPotQA training set to construct session-level trajectories. Each trajectory begins with the \texttt{[GetUserQuestion]} prompt. We then repeatedly prompt GPT-4 to predict actions between \texttt{[Search]} and \texttt{[PredictAnswer]}. If GPT-4 predicts \texttt{[Search]}, we prompt it to generate a search query and append the corresponding search results to the trajectory, continuing this cycle. This process continues until GPT-4 predicts \texttt{[PredictAnswer]}. If the answer is correct (as evaluated by the GPT-4 evaluator), we replace the predicted answer with the ground-truth answer; otherwise, the data is discarded. Additionally, if GPT-4 predicts \texttt{[Search]} five times in a session, we terminate and discard the data. 

Next, for each trajectory, where there are $k$ rounds, we prompt GPT-3.5 using the first $k-1$ rounds as context to decide the final round's action: \texttt{[PredictAnswer]} or \texttt{[SeekAdvice]}. If GPT-3.5 selects \texttt{[SeekAdvice]}, we replace the final step with \texttt{[SeekAdvice]} and the corresponding thoughts from GPT-3.5. Otherwise, the original trajectory remains unchanged.

This process results in 10,240 session-level trajectories for the imitation learning stage. For the reinforcement learning stage, we directly use the original HotPotQA training set, consisting of 90,447 samples.

\subsection{Defining proxy reward for RL}\label{state-value-function}

In the question-answering tasks, sessions are not independent. Actions taken in earlier sessions can influence memory, creating lasting effects on subsequent sessions. As illustrated in Equation \eqref{reward-0}, the term $A_i := V_{\pi_{\theta_k}}(\mathcal{S}_{i+1}) - V_{\pi_{\theta_k}}(\mathcal{S}_i)$ measures the advantage of state $\mathcal{S}_{i+1}$ over state $\mathcal{S}_i$ (note that $\mathcal{S}_i$ here represents the initial state of session $i$). In our experiment setting, if the agent predicts \texttt{[SeekAdvice]}, it will receive expert advice, extract some knowledge by reflection, and write that knowledge to the memory. Intuitively, $A_i$ should increase if the new knowledge is useful in subsequent sessions, and it should decrease if there is already a lot of similar knowledge in the memory at the start of session $i$. Hence, we use the following heuristic definition,
\begin{equation}
    A_i = \beta\frac{\mathbbm{I}(N_{i+1:n}(q_i) > 0)}{M_{1:i-1}(q_i) + 1},
\label{advantage_reward}
\end{equation}
where $q_i$ represents the user question in session $i$; $N_{i+1:n}(q_i)$ represents the number of user questions in session $i+1$ to session $n$ that are similar enough to $q_i$; $M_{0:i-1}(q_i)$ represents the number of user questions in session $1$ to session $i-1$ that are both similar enough to $q_i$ and added to the memory; $\mathbbm{I}(\cdot)$ is the indicator function. $\beta$ is a hyperparameter, we set $\beta = 0.1$ by default.

\section{Supplementary experimental results on RL training}\label{ppo_training}

In this section, we present detailed experimental results for RL training on ProductQA.

\subsection{Training curve}

\begin{figure}[ht]
    \centering
    \includegraphics[width=1.0\linewidth]{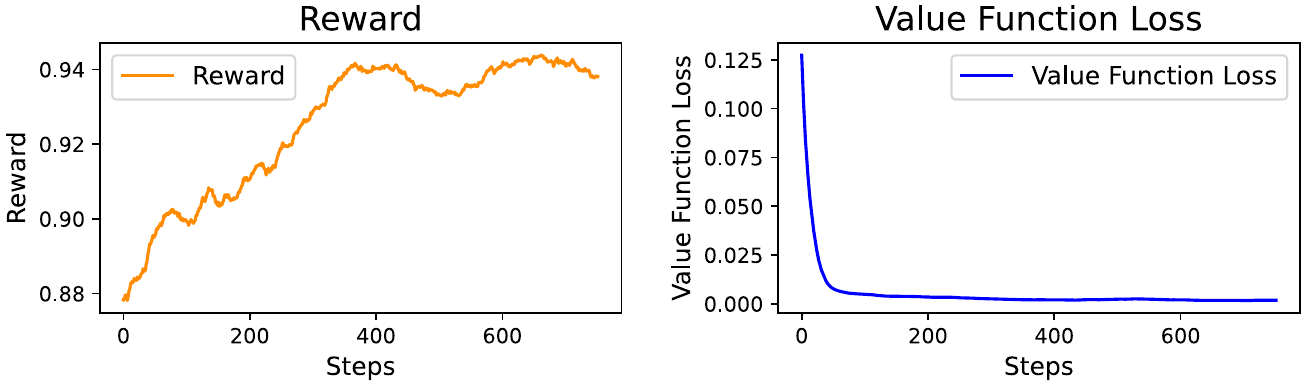}
    \caption{Reward and value function loss curves during the PPO training process on ProductQA.}
    \label{fig:loss-and-reward}
\end{figure}

In Figure~\ref{fig:loss-and-reward}, we provide training curves, indicating that RL training converged after 500 steps.

\subsection{Training Robustness}

We conduct multiple independent trials of PPO training to study the variation of the result, as shown in Table~\ref{robustness}. On average, RL training improves the total score by 2.6\%, with a standard deviation of 0.3\%, demonstrating the significance of RL improvements.

\begin{table}[ht]
\renewcommand{\arraystretch}{1}
\centering
\caption{Robustness of RL training. Here, w/o RL represents the agent trained solely by imitation learning. \texttt{agile-vic13b-ppo-X} stands for the X-th RL experiment. The table presents the average and standard deviation across multiple RL training runs.}\label{robustness}
\begin{tabular}{lcccc}
\toprule[1pt]
\multicolumn{1}{l}{\multirow{2}{*}{\textbf{Method}}} & \multirow{2}{*}{\textbf{Advice Rate}~$\downarrow$} & \multirow{2}{*}{\textbf{Accuracy}~$\uparrow$} & \multirow{2}{*}{\textbf{Total Score}~$\uparrow$} & \textbf{Relative Improvement} \\
 &  &  &  & \textbf{to w/o RL} \\
\midrule[0.5pt]
w/o RL & 0.256 & 0.843 & 0.766 & - \\
\midrule[0.5pt]
agile-vic13b-ppo-1 & 0.233 & 0.854 & 0.784 & 2.3\% \\
agile-vic13b-ppo-2 & 0.226 & 0.855 & 0.787 & 2.7\% \\
agile-vic13b-ppo-3 & 0.209 & 0.851 & 0.788 & 2.9\% \\
\midrule[0.5pt]
average & 0.223 & 0.853 & 0.786 & 2.6\% \\
standard deviation & 0.012 & 0.002 & 0.002 & 0.3\% \\
\bottomrule[1pt]
\end{tabular}
\end{table}

\subsection{Impact of PPO training}

To further investigate the impact of PPO training in more general and varied scenarios, we conducted additional experiments in two distinct settings.

First, we re-generated SFT training data for \texttt{agile-vic13b-sft} such that the agent performs \texttt{[SeekAdvice]} randomly in 25\% of cases. This initial policy is simpler but more general. In this setting, we name the SFT model \texttt{agile-vic13b-sft-random}, and the final model trained with RL on top of it \texttt{agile-vic13b-ppo-random}. As shown in Table~\ref{ppo_improvement}, RL training brings a 7.1\% improvement in this setting. Interestingly, the performance of \texttt{agile-vic13b-ppo-random} is better than that of \texttt{agile-vic13b-ppo}. We conjecture that random seeking-advice is a better initial policy because it enables exploration in all directions.

In the second experiment, we lowered the advice cost to 0.1. After PPO training, as shown in Table~\ref{ppo_improvement}, the \texttt{agile-vic13b-ppo-random} agent quickly adapted to the new cost, performing \texttt{[SeekAdvice]} much more aggressively than the initial agent trained by SFT. In this scenario, RL training brings a 22.3\% improvement.

\begin{table}[ht]
\renewcommand{\arraystretch}{1}
\centering
\caption{Improvement of PPO training. The training data for \texttt{agile-vic13b-sft} includes trajectories from GPT-4 agent. The training data for \texttt{agile-vic13b-random} is constructed by randomly assigning \texttt{[SeekAdvice]} to 25\% of the data. \texttt{agile-vic13b-ppo} and \texttt{agile-vic13b-ppo-random} are initialized from \texttt{agile-vic13b-sft}  and \texttt{agile-vic13b-sft-random}, respectively, and both are trained with PPO.}
\label{ppo_improvement}
\begin{tabular}{lcccc}
\toprule[1pt]
\textbf{Method} & \textbf{seeking advice cost} & \textbf{Advice Rate}~$\downarrow$ & \textbf{Accuracy}~$\uparrow$ & \textbf{Total Score}~$\uparrow$  \\
\midrule[0.5pt]
agile-vic13b-sft & 0.3 & 0.256 & 0.843 & 0.766 \\
agile-vic13b-ppo & 0.3 & 0.233 & 0.854 & 0.784{\tiny{(+2.3\%)}} \\
\midrule[0.5pt]
agile-vic13b-sft-random & 0.3 & 0.014 & 0.749 & 0.745 \\
agile-vic13b-ppo-random & 0.3 & 0.306 & 0.89 & 0.798{\tiny{(+7.1\%)}} \\
\midrule[0.5pt]
agile-vic13b-sft-random & 0.1 & 0.014 & 0.749 & 0.748 \\
agile-vic13b-ppo-random & 0.1 & 0.671 & 0.981 & 0.914{\tiny{(+22.3\%)}} \\
\bottomrule[1pt]
\end{tabular}
\end{table}

\newpage
\section{Tables}\label{sec:Tables}

\begin{table}[ht]
\renewcommand{\arraystretch}{1}
\centering
\caption{Statistics of the ProductQA dataset. \# Products indicates the number of products within each group. \# Fact-QA, \# Search-QA and \# Reasoning-QA display the respective numbers of QA pairs categorized as Fact-QA, Search-QA, and Reasoning-QA.}
\label{data_statistic}
\vspace*{\fill}
\resizebox{0.99\textwidth}{!}{
\begin{tabular}{llccccc}
\toprule[1pt]
\multicolumn{2}{c}{\textbf{Groups}} & \textbf{\# Products} & \textbf{\# Fact-QA} & \textbf{\# Search-QA} & \textbf{\# Reasoning-QA} & \textbf{Total} \\
\midrule[0.5pt]
\multirow{21}{*}{\makecell[l]{Train}}
 & Blades & 20 & 2,147 & 769 & 631 & 3,547 \\
 & Headlight Bulbs & 20 & 1,767 & 644 & 463 & 2,874 \\
 & Cell Phones & 20 & 1,636 & 761 & 374 & 2,771 \\
 & Portable Power Banks & 20 & 3,344 & 673 & 500 & 4,517 \\
 & Dresses & 20 & 2,287 & 738 & 263 & 3,288 \\
 & Everyday Bras & 20 & 1,942 & 684 & 336 & 2,962 \\
 & Wrist Watches & 20 & 2,169 & 757 & 389 & 3,315 \\
 & Blu-ray Players & 20 & 1,630 & 688 & 572 & 2,890 \\
 & Camera Lenses & 20 & 1,859 & 769 & 1,025 & 3,653 \\
 & Headphones & 20 & 5,432 & 766 & 583 & 6,781 \\
 & Mice & 20 & 5,653 & 490 & 294 & 6,437 \\
 & Point \& Shoot Digital Cameras & 20 & 1,696 & 722 & 565 & 2,983 \\
 & Coffee Machines & 20 & 4,184 & 681 & 638 & 5,503 \\
 & Digital Scales & 20 & 2,724 & 391 & 682 & 3,797 \\
 & Space Heaters & 20 & 2,283 & 674 & 498 & 3,455 \\
 & Printers & 20 & 1,431 & 760 & 489 & 2,680 \\
 & Litter & 20 & 1,860 & 753 & 507 & 3,120 \\
 & Grips & 20 & 1,771 & 713 & 413 & 2,897 \\
 & Gun Holsters & 20 & 1,679 & 94 & 1,362 & 3,135 \\
 & Handheld Flashlights & 20 & 2,009 & 768 & 482 & 3,259 \\
 \cmidrule[0.5pt](r){2-7}
 & Total & 400 & 49,503 & 13,295 & 11,066 & 73,864 \\
\midrule[0.5pt]
\multirow{7}{*}{\makecell[l]{Test}}
 & Leggings & 20 & 969 & 743 & 527 & 2,239 \\
 & Camera Cases & 20 & 975 & 706 & 898 & 2,579 \\
 & Motherboards & 20 & 989 & 736 & 826 & 2,551 \\
 & All Pans & 20 & 973 & 747 & 275 & 1,995 \\
 & Rollerball Pens & 20 & 967 & 760 & 603 & 2,330 \\
 & Rifle Scopes & 17 & 979 & 714 & 978 & 2,671 \\
 \cmidrule[0.5pt](r){2-7}
 & Total & 117 & 5,852 & 4,406 & 4,107 & 14,365 \\
\bottomrule[1pt]
\end{tabular}
}
\vspace*{\fill}
\end{table}

\begin{table}[ht]
\renewcommand{\arraystretch}{1}
\centering
\caption{Training statistics for each experiment.}
\label{training_times}
\begin{tabular}{lccc}
\toprule[1pt]
\multicolumn{1}{l}{\textbf{Task}} & \textbf{Number of H800 GPU} & \textbf{SFT Training Time} & \textbf{RL Training Time} \\
\midrule[0.5pt]
ProductQA & 8 & 3.6 hours & 5.5 hours \\
\midrule[0.5pt]
MedMCQA & 8 & 0.9 hours & 2.0 hours \\
\midrule[0.5pt]
HotPotQA & 8 & 7.9 hours & 27.5 hours \\
\bottomrule[1pt]
\end{tabular}
\end{table}

\begin{landscape}
\begin{table}[t]
\renewcommand{\arraystretch}{1}
\centering
\caption{Detail performance of our methods and other baselines on six test product groups of ProductQA. \texttt{X-prompt} represents directly prompting the model X; \texttt{agile-X-Y} represents incorporating the model X within the AGILE framework, while Y represents prompting or PPO training. The Short and Long stand for the results evaluated on short answers and long answers, respectively. The seeking advice cost is $c=0.3$. The best total scores are highlighted in bold.}
\label{overall_result_appendix}
\scalebox{1}{
\begin{tabular}{llcccccccccccccc}
\toprule[1pt]
\multicolumn{2}{c}{\multirow{3}{*}[-0.9ex]{\textbf{Group}}} & \multicolumn{2}{c}{\textbf{gpt3.5-}} & \multicolumn{2}{c}{\textbf{gpt4-}} & \multicolumn{2}{c}{\textbf{agile-vicuna-}} & \multicolumn{2}{c}{\textbf{agile-gpt3.5-}} & \multicolumn{2}{c}{\textbf{agile-gpt4-}} & \multicolumn{2}{c}{\textbf{agile-vic7b-}} & \multicolumn{2}{c}{\textbf{agile-vic13b-}} \\
&& \multicolumn{2}{c}{\textbf{prompt}} & \multicolumn{2}{c}{\textbf{prompt}} & \multicolumn{2}{c}{\textbf{13b-prompt}} & \multicolumn{2}{c}{\textbf{prompt}} & \multicolumn{2}{c}{\textbf{prompt}} & \multicolumn{2}{c}{\textbf{ppo(ours)}} & \multicolumn{2}{c}{\textbf{ppo(ours)}} \\
\cmidrule[0.5pt](r){3-4} \cmidrule[0.5pt](r){5-6} \cmidrule[0.5pt](r){7-8} \cmidrule[0.5pt](r){9-10} \cmidrule[0.5pt](r){11-12} \cmidrule[0.5pt](r){13-14} \cmidrule[0.5pt](r){15-16}
&& \textbf{Short} & \textbf{Long} & \textbf{Short} & \textbf{Long} & \textbf{Short} & \textbf{Long} & \textbf{Short} & \textbf{Long} & \textbf{Short} & \textbf{Long} & \textbf{Short} & \textbf{Long} & \textbf{Short} & \textbf{Long} \\
\midrule[0.5pt]
\multirow{3}{*}{\makecell[l]
{Camera\\Cases}}
& Advice Rate~$\downarrow$ & - & - & - & - & 0.182 & 0.182 & 0.313 & 0.313 & 0.175 & 0.175 & 0.199 & 0.199 & 0.263 & 0.263 \\
& Accuracy~$\uparrow$ & 0.200 & 0.320 & 0.385 & 0.495 & 0.182 & 0.330 & 0.537 & 0.644 & 0.775 & 0.791 & 0.818 & 0.776 & 0.860 & 0.841 \\
& Total Score~$\uparrow$ & - & - & - & - & 0.127 & 0.275 & 0.443 & 0.550 & 0.722 & 0.738 & 0.758 & 0.716 & \textbf{0.781} & \textbf{0.762} \\
\midrule[0.5pt]
\multirow{3}{*}{Leggings}
& Advice Rate~$\downarrow$ & - & - & - & - & 0.154 & 0.154 & 0.359 & 0.359 & 0.200 & 0.200 & 0.201 & 0.201 & 0.251 & 0.251 \\
& Accuracy~$\uparrow$ & 0.181 & 0.306 & 0.503 & 0.594 & 0.154 & 0.267 & 0.497 & 0.646 & 0.766 & 0.790 & 0.837 & 0.834 & 0.876 & 0.885 \\
& Total Score~$\uparrow$ & - & - & - & - & 0.108 & 0.221 & 0.389 & 0.538 & 0.706 & 0.730 & 0.777 & 0.774 & \textbf{0.801} & \textbf{0.810} \\
\midrule[0.5pt]
\multirow{3}{*}{\makecell[l]
{All\\Pans}}
& Advice Rate~$\downarrow$ & - & - & - & - & 0.167 & 0.167 & 0.336 & 0.336 & 0.220 & 0.220 & 0.184 & 0.184 & 0.220 & 0.220 \\
& Accuracy~$\uparrow$ & 0.201 & 0.297 & 0.470 & 0.538 & 0.167 & 0.272 & 0.506 & 0.605 & 0.784 & 0.804 & 0.843 & 0.831 & 0.866 & 0.869 \\
& Total Score~$\uparrow$ & - & - & - & - & 0.117 & 0.222 & 0.405 & 0.504 & 0.718 & 0.738 & 0.788 & 0.776 & \textbf{0.800} & \textbf{0.803} \\
\midrule[0.5pt]
\multirow{3}{*}{\makecell[l]
{Rollerball\\Pens}}
& Advice Rate~$\downarrow$ & - & - & - & - & 0.130 & 0.130 & 0.333 & 0.333 & 0.231 & 0.231 & 0.162 & 0.162 & 0.212 & 0.212 \\
& Accuracy~$\uparrow$ & 0.193 & 0.271 & 0.449 & 0.573 & 0.130 & 0.242 & 0.482 & 0.627 & 0.767 & 0.808 & 0.776 & 0.769 & 0.816 & 0.824 \\
& Total Score~$\uparrow$ & - & - & - & - & 0.091 & 0.203 & 0.382 & 0.527 & 0.698 & 0.739 & 0.727 & 0.720 & \textbf{0.752} & \textbf{0.760} \\
\midrule[0.5pt]
\multirow{3}{*}{\makecell[l]
{Mother-\\boards}}
& Advice Rate~$\downarrow$ & - & - & - & - & 0.214 & 0.214 & 0.303 & 0.303 & 0.225 & 0.225 & 0.162 & 0.162 & 0.235 & 0.235 \\
& Accuracy~$\uparrow$ & 0.253 & 0.431 & 0.511 & 0.637 & 0.215 & 0.337 & 0.525 & 0.686 & 0.815 & 0.855 & 0.835 & 0.831 & 0.877 & 0.882 \\
& Total Score~$\uparrow$ & - & - & - & - & 0.151 & 0.273 & 0.434 & 0.595 & 0.747 & 0.788 & 0.786 & 0.782 & \textbf{0.806} & \textbf{0.812} \\
\midrule[0.5pt]
\multirow{3}{*}{\makecell[l]
{Rifle\\Scopes}}
& Advice Rate~$\downarrow$ & - & - & - & - & 0.197 & 0.197 & 0.293 & 0.293 & 0.198 & 0.198 & 0.167 & 0.167 & 0.216 & 0.216 \\
& Accuracy~$\uparrow$ & 0.187 & 0.306 & 0.463 & 0.587 & 0.197 & 0.313 & 0.502 & 0.657 & 0.770 & 0.806 & 0.802 & 0.760 & 0.828 & 0.822 \\
& Total Score~$\uparrow$ & - & - & - & - & 0.138 & 0.254 & 0.414 & 0.569 & 0.711 & 0.747 & 0.752 & 0.710 & \textbf{0.763} & \textbf{0.757} \\
\midrule[0.5pt]
\multirow{3}{*}{Average}
& Advice Rate~$\downarrow$ & - & - & - & - & 0.174 & 0.174 & 0.323 & 0.323 & 0.208 & 0.208 & 0.179 & 0.179 & 0.233 & 0.233 \\
& Accuracy~$\uparrow$ & 0.202 & 0.322 & 0.464 & 0.571 & 0.174 & 0.294 & 0.508 & 0.644 & 0.780 & 0.809 & 0.818 & 0.800 & 0.854 & 0.854 \\
& Total Score~$\uparrow$ & - & - & - & - & 0.122 & 0.242 & 0.411 & 0.547 & 0.718 & 0.747 & 0.764 & 0.746 & \textbf{0.784} & \textbf{0.784} \\
\bottomrule[1pt]
\end{tabular}
}
\vspace{-0.2cm}
\end{table}
\end{landscape}
\newpage

\begin{table}[htbp]
\renewcommand{\arraystretch}{1}
\centering
\caption{Ablation study on ProductQA test tasks. {\bf w/o Reflection} represents removing the reflection function. {\bf w/o Memory} represents prohibiting memory component. {\bf w/o Advice} represents removing the seeking advice function. {\bf Non-adapt advice} represents seeking advice in the same number with \texttt{agile-vic13b-ppo} at the beginning of trajectory. {\bf w/o Tool-Use} represents removing the search product function. {\bf w/o RL} represents the \texttt{agile-vic13b-sft}. The best scores are highlighted in bold.}
\label{ablation_appendix}
\scalebox{0.78}{
\begin{tabular}{llccccccc}
\toprule[1pt]
\multicolumn{2}{c}{\multirow{2}{*}{\textbf{Group}}} & \textbf{w/o} & \textbf{w/o} & \textbf{w/o} & \textbf{Non-adapt} & \textbf{w/o} & \textbf{w/o} & \textbf{agile-vic-} \\
&& \textbf{Reflection} & \textbf{Memory} & \textbf{Advice} & \textbf{Advice} & \textbf{Tool-Use} & \textbf{RL} & \textbf{13b-ppo} \\
\midrule[0.5pt]
\multirow{3}{*}{\makecell[l]
{Camera\\Cases}}
& Advice Rate~$\downarrow$ & 0.335 & 0.459 & 0.000 & 0.263 & 0.452 & 0.295 & 0.263 \\
& Accuracy~$\uparrow$ & 0.851 & 0.869 & 0.735 & 0.827 & 0.870 & 0.849 & 0.860 \\
& Total Score~$\uparrow$ & 0.750{\tiny{(-4.1\%)}} & 0.731{\tiny{(-6.8\%)}} & 0.735{\tiny{(-6.3\%)}} & 0.748{\tiny{(-4.4\%)}} & 0.734{\tiny{(-6.4\%)}} & 0.760{\tiny{(-2.8\%)}} & \textbf{0.781} \\
\midrule[0.5pt]
\multirow{3}{*}{Leggings}
& Advice Rate~$\downarrow$ & 0.276 & 0.437 & 0.000 & 0.251 & 0.529 & 0.290 & 0.251 \\
& Accuracy~$\uparrow$ & 0.874 & 0.902 & 0.762 & 0.828 & 0.880 & 0.867 & 0.876 \\
& Total Score~$\uparrow$ & 0.791{\tiny{(-1.3\%)}} & 0.771{\tiny{(-3.9\%)}} & 0.762{\tiny{(-5.1\%)}} & 0.753{\tiny{(-6.4\%)}} & 0.721{\tiny{(-11.1\%)}} & 0.780{\tiny{(-2.7\%)}} & \textbf{0.801} \\
\midrule[0.5pt]
\multirow{3}{*}{All Pans}
& Advice Rate~$\downarrow$ & 0.263 & 0.413 & 0.000 & 0.220 & 0.550 & 0.225 & 0.220 \\
& Accuracy~$\uparrow$ & 0.867 & 0.900 & 0.759 & 0.818 & 0.877 & 0.855 & 0.866 \\
& Total Score~$\uparrow$ & 0.788{\tiny{(-1.5\%)}} & 0.776{\tiny{(-3.1\%)}} & 0.759{\tiny{(-5.4\%)}} & 0.752{\tiny{(-6.4\%)}} & 0.712{\tiny{(-12.4\%)}} & 0.788{\tiny{(-1.5\%)}} & \textbf{0.800} \\
\midrule[0.5pt]
\multirow{3}{*}{\makecell[l]
{Rollerball\\Pens}}
& Advice Rate~$\downarrow$ & 0.237 & 0.378 & 0.000 & 0.212 & 0.501 & 0.220 & 0.212 \\
& Accuracy~$\uparrow$ & 0.818 & 0.843 & 0.727 & 0.785 & 0.868 & 0.812 & 0.816 \\
& Total Score~$\uparrow$ & 0.747{\tiny{(-0.7\%)}} & 0.730{\tiny{(-3.0\%)}} & 0.727{\tiny{(-3.4\%)}} & 0.721{\tiny{(-4.3\%)}} & 0.718{\tiny{(-4.7\%)}} & 0.746{\tiny{(-0.8\%)}} & \textbf{0.752} \\
\midrule[0.5pt]
\multirow{3}{*}{\makecell[l]
{Mother-\\boards}}
& Advice Rate~$\downarrow$ & 0.270 & 0.368 & 0.000 & 0.235 & 0.483 & 0.285 & 0.235 \\
& Accuracy~$\uparrow$ & 0.878 & 0.886 & 0.766 & 0.829 & 0.873 & 0.871 & 0.877 \\
& Total Score~$\uparrow$ & 0.797{\tiny{(-1.1\%)}} & 0.776{\tiny{(-3.9\%)}} & 0.766{\tiny{(-5.2\%)}} & 0.758{\tiny{(-6.3\%)}} & 0.728{\tiny{(-10.7\%)}} & 0.786{\tiny{(-2.5\%)}} & \textbf{0.806} \\
\midrule[0.5pt]
\multirow{3}{*}{\makecell[l]
{Rifle\\Scopes}}
& Advice Rate~$\downarrow$ & 0.237 & 0.385 & 0.000 & 0.216 & 0.440 & 0.221 & 0.216 \\
& Accuracy~$\uparrow$ & 0.824 & 0.858 & 0.733 & 0.783 & 0.824 & 0.805 & 0.828 \\
& Total Score~$\uparrow$ & 0.753{\tiny{(-1.3\%)}} & 0.742{\tiny{(-2.8\%)}} & 0.733{\tiny{(-4.1\%)}} & 0.718{\tiny{(-6.3\%)}} & 0.692{\tiny{(-10.3\%)}} & 0.739{\tiny{(-3.2\%)}} & \textbf{0.763} \\
\midrule[0.5pt]
\multirow{3}{*}{Average}
& Advice Rate~$\downarrow$ & 0.270 & 0.407 & 0.000 & 0.233 & 0.492 & 0.256 & 0.233 \\
& Accuracy~$\uparrow$ & 0.852 & 0.876 & 0.747 & 0.812 & 0.865 & 0.843 & 0.854 \\
& Total Score~$\uparrow$ & 0.771{\tiny{(-1.7\%)}} & 0.754{\tiny{(-4.0\%)}} & 0.747{\tiny{(-5.0\%)}} & 0.742{\tiny{(-5.7\%)}} & 0.717{\tiny{(-9.3\%)}} & 0.766{\tiny{(-2.3\%)}} & \textbf{0.784} \\
\bottomrule[1pt]
\end{tabular}
}
\vspace{-0.2cm}
\end{table}

\begin{table}[ht]
\renewcommand{\arraystretch}{1}
\centering
\caption{Performance of the model (\texttt{agile-vic13b-ppo}) trained on different seeking advice cost settings.}
\label{ablation_reward}
\begin{tabular}{llccccc}
\toprule[1pt]
\multicolumn{2}{c}{\multirow{2}{*}[-0.9ex]{\textbf{Group}}} & \multicolumn{5}{c}{\textbf{Seeking Advice Cost}} \\
\cmidrule[0.5pt](r){3-7}
&& \textbf{0.5} & \textbf{0.4} & \textbf{0.3} & \textbf{0.2} & \textbf{0.1}\\
\midrule[0.5pt]
\multirow{2}{*}{Camera Cases}
& Advice Rate & 0.108 & 0.189 & 0.263 & 0.339 & 0.458 \\
& Accuracy & 0.806 & 0.829 & 0.860 & 0.885 & 0.929 \\
\midrule[0.5pt]
\multirow{2}{*}{Leggings}
& Advice Rate & 0.098 & 0.188 & 0.251 & 0.317 & 0.464 \\
& Accuracy & 0.824 & 0.844 & 0.876 & 0.877 & 0.921 \\
\midrule[0.5pt]
\multirow{2}{*}{All Pans}
& Advice Rate & 0.094 & 0.163 & 0.220 & 0.262 & 0.384 \\
& Accuracy & 0.813 & 0.845 & 0.866 & 0.889 & 0.926 \\
\midrule[0.5pt]
\multirow{2}{*}{Rollerball Pens}
& Advice Rate & 0.100 & 0.163 & 0.212 & 0.264 & 0.406 \\
& Accuracy & 0.780 & 0.799 & 0.816 & 0.829 & 0.891 \\
\midrule[0.5pt]
\multirow{2}{*}{Motherboards}
& Advice Rate & 0.103 & 0.162 & 0.235 & 0.307 & 0.443 \\
& Accuracy & 0.825 & 0.839 & 0.877 & 0.901 & 0.941 \\
\midrule[0.5pt]
\multirow{2}{*}{Rifle Scopes}
& Advice Rate & 0.087 & 0.144 & 0.216 & 0.257 & 0.385 \\
& Accuracy & 0.780 & 0.797 & 0.828 & 0.845 & 0.897 \\
\midrule[0.5pt]
\multirow{2}{*}{Average}
& Advice Rate & 0.098 & 0.168 & 0.233 & 0.291 & 0.423 \\
& Accuracy & 0.805 & 0.825 & 0.854 & 0.871 & 0.918 \\
\bottomrule[1pt]
\end{tabular}
\vspace{-0.2cm}
\end{table}

\newpage

\section{Case study}\label{app:case_study}

Case \#1, illustrated in Table~\ref{tab:case_study_1}, provides a specific example demonstrating how \texttt{agile-vic13b-ppo} proactively seeks advice from a human expert for questions it cannot answer. Furthermore, it leverages reflection to extract general knowledge from the expert's responses, which can then be applied in future QA sessions.

Case \#2, shown in Table~\ref{tab:case_study_2}, demonstrates how \texttt{agile-vic13b-ppo} utilizes tools to address product recommendation questions.

\begin{landscape}
\begin{table}[ht]
    \centering
    \caption{Case \#1 from \texttt{agile-vic13b-ppo} on the test task, camera cases group, in ProductQA. \myCheckMark represents correct responses. Our agent proactively seeks advice from human experts at session 313 and summarizes a piece of general knowledge by reflection. In the following sessions, the agent leverages this knowledge to successfully answer three questions for different products at sessions 721, 1265, and 1364, respectively. It demonstrates the effectiveness of seeking advice and reflection.}
    \label{tab:case_study_1}
    \scalebox{0.97}{
        \begin{tabular}{cccm{6.4cm}}
\toprule[1pt]
            \multicolumn{1}{c}{\textbf{Session}} & \multicolumn{1}{c}{\textbf{Context}} & \multicolumn{1}{c}{\textbf{Action}} & \multicolumn{1}{c}{\textbf{Reflection / Answer}}\\
\midrule[0.5pt]

\multicolumn{1}{c}{\multirow{6}{*}{313}} & \makecell[l]{Answer users' [Question] about product B00PV0IEDY based on the following three \\ types of information:\\ 1. \textbf{[Product Metadata]} ... `Carrying Options': `backpack straps' ... \\ 2. \textbf{[Memory]} Existing knowledge in your memory about camera\_cases.\\ camera cases with no tripod mount are not suitable for scenarios where a stable camera} & \multirow{6}{*}{\makecell[c]{Seek\\Advice}} & \textbf{Human Expert:} Certainly, the backpack straps on the camera case are designed for comfort even when carrying heavier loads. These straps enable ergonomic weight distribution, making the case ideal for prolonged travel. \myCheckMark \\
& \makecell[l]{setup is crucial, such as long-exposure photography or macro work where camera shake \\ can significantly impact image quality. \quad 3.\textbf{[History]} ... \\ ... \\ \textbf{Question}: Is the camera case designed for prolonged travel and capable of withstanding  \\ heavy loads? } && \textbf{Reflection:} \textsl{\uline{Camera cases with backpack straps feature ergonomic designs that help distribute the weight of the camera equipment across the user's back, reducing strain and increasing comfort during transport.}} \\
\midrule[0.5pt]
\multicolumn{1}{c}{721} & \makecell[l]{Answer users' [Question] about product B000WH86BQ ...\\ 1. \textbf{[Product Metadata]} ... `Carrying Options': `backpack straps' ... \\ 2. \textbf{[Memory]} Existing knowledge in your memory about camera\_cases.\\ \textsl{\underline{Camera cases with backpack straps feature ergonomic designs that help distribute the}} \\ \textsl{\underline{weight of the camera equipment across the user's back ...}} \quad 3.\textbf{[History]} ... \\ \textbf{Question}: Is the camera case suitable for extended journeys and potentially heavy loads?}  & \makecell[c]{Predict\\Answer} & \textbf{Answer:} The case is designed to be adjustable and has padded straps which makes it suitable for long-term use with heavy equipment. \myCheckMark \\
\midrule[0.5pt]
\multicolumn{1}{c}{1265} & \makecell[l]{Answer users' [Question] about product B005J09OXU ...\\ 1. \textbf{[Product Metadata]} ... `Carrying Options': `backpack straps' ... \\ 2. \textbf{[Memory]} Existing knowledge in your memory about camera\_cases.\\ \textsl{\underline{Camera cases with backpack straps feature ergonomic designs that help distribute the}} \\ \textsl{\underline{weight of the camera equipment across the user's back ...}} \quad 3.\textbf{[History]} ... \\ \textbf{Question}: Will this camera case enhance the comfort of transporting my gear on treks?}  & \makecell[c]{Predict\\Answer} &\textbf{Answer:} Certainly, a camera case with backpack straps is engineered to evenly distribute the weight across your shoulders, improving ergonomics and comfort. This is particularly beneficial for reducing fatigue and strain during prolonged activities like treks. \myCheckMark \\
\midrule[0.5pt]
\multicolumn{1}{c}{1364} & \makecell[l]{Answer users' [Question] about product B000YA33DC ...\\ 1. \textbf{[Product Metadata]} ... `Carrying Options': `backpack straps' ... \\ 2. \textbf{[Memory]} Existing knowledge in your memory about camera\_cases.\\ \textsl{\underline{Camera cases with backpack straps feature ergonomic designs that help distribute the}} \\ \textsl{\underline{weight of the camera equipment across the user's back ...}} \quad 3.\textbf{[History]} ... \\ \textbf{Question}: Does this camera case facilitate easier transportation when I frequently travel?}  & \makecell[c]{Predict\\Answer} & \textbf{Answer:} Certainly, the backpack straps on this camera case not only simplify the process of carrying but also evenly distribute the weight, making it an ideal option for those who are frequently on the move and prefer a more comfortable and balanced carrying solution.
 \myCheckMark \\
\bottomrule[1pt]
        \end{tabular}
    }
    \label{case_study_seek_advice}
\end{table}

\begin{table}[t]
    \centering
    \caption{Case \#2 from \texttt{agile-vic13b-ppo} on the ProductQA test set. \myCheckMark represents correct responses. By leveraging SQL searching tools, agent successfully find the correct product to recommend to users.}
    \label{tab:case_study_2}
    \scalebox{1}{
        \begin{tabular}{cccm{5cm}}
\toprule[1pt]
            \multicolumn{1}{c}{\textbf{Session}} & \multicolumn{1}{c}{\textbf{Context}} & \multicolumn{1}{c}{\textbf{Action}} & \multicolumn{1}{c}{\textbf{Output}}\\
\midrule[0.5pt]
918 & \makecell[l]{Answer users' [Question] about product B00FUHWE5M based on the following three \\ types of information:\\ 1. \textbf{[Product Metadata]} provides metadata about product B00FUHWE5M \\ from the leggings database. The schema of the leggings database is outlined below \\ (in the format field[unit](value1, value2, ...)). \\ price[dollar] \\ \textsl{\underline{brand}}(Jescakoo, QZUnique, ...) \\ warmth(hot,moderate,cool) ... \\ 2. \textbf{[Memory]} ... \\ 3. \textbf{[History]} ... \\ ... \\ \textbf{Question}: I've heard good things about \underline{QZUnique}. Spit out the first leggings you have \\ from them. ... } & \makecell[c]{Search\\Product} & \textbf{SQL:} SELECT product\_id FROM ``leggings'' WHERE \textsl{\underline{brand}} = ``\underline{QZUnique}'' LIMIT 1;\newline\newline \textbf{Answer:} The product that meets your requirements is as follows. `QZUnique Women's Classic Retro Printed Casual Pattern Ankle Length Elastic Tights Leggings' with asin: B012IKGF62 \myCheckMark \\
\midrule[0.5pt]
1298 & \makecell[l]{Answer users' [Question] about product B012AQGKXC based on the following three types \\ of information:\\ 1. \textbf{[Product Metadata]} provides metadata about product B012AQGKXC \\ from the motherboards database. The schema of the motherboards database is outlined below \\ (in the format field[unit](value1, value2, ...)). \\ price[dollar] \\ \textsl{\underline{brand}}(ASRock, Gigabyte, ...) \\ \textsl{\underline{memory\_support}}[GB] ... \\ 2. \textbf{[Memory]} ... \\ 3. \textbf{[History]} ... \\ ... \\ \textbf{Question}: I'm looking for an \underline{ASRock} motherboard with at least \underline{32 GB} of memory support.  \\ Don't keep me waiting, hustle up and find it for me. ... } & \makecell[c]{Search\\Product} & \textbf{SQL:} SELECT product\_id FROM ``motherboards'' WHERE \underline{brand} = ``\underline{ASRock}'' AND \textsl{\underline{memory\_support}} >= \underline{32} LIMIT 1; \newline\newline \textbf{Answer:} The product that would suit your preferences is: `AS Rock LGA1155 DDR3 SATA3 USB3.0 Quad  CrossFireX and Quad SLI A E ATX Motherboard Z77 EXTREME4' with asin: B007KTY4A6 \myCheckMark \\

\bottomrule[1pt]
        \end{tabular}
    }
    \label{case_study_search_product}
\end{table}

\end{landscape}

\newpage
\section{Development of the ProductQA dataset}\label{sec:Development of the ProductQA dataset}

\subsection{Product collection}~\label{product_collection}
The product groups and the corresponding products are collected by the following steps.

\begin{enumerate}
    \item Filter the Amazon Review Data to retain only products with at least 100 reviews, then cluster them by category tags.
    \item Sort the clusters by size, from largest to smallest. Manually review each cluster in order: we keep product clusters that involve diverse technical details and long-tail domain knowledge, such as electronics, from which we can potentially construct a diverse set of user questions. The manual review ends when we have collected 26 clusters. Each cluster is referred to as a \emph{product group}.
    \item For each product group, we remove the top 10\% of products with the highest number of reviews. We exclude these most popular products from the datasets to prevent data leakage, as information about them is likely included in the pre-training set of LLMs. From the remaining items, we randomly select up to 20 products to form the final product set.
\end{enumerate}

\subsection{Annotation guidelines}~\label{annotation_guidelines}
There are two annotation tasks, product table creation and QA collection. We provide the annotation guidelines in this Section.

\paragraph{Task 1: Product table creation}
For each product group, we provide a series of features and their corresponding values for each product in the group. This information is obtained by prompting GPT-4 to extract data from the reviews of each product. The task of annotators is to construct a product table containing only the metadata. Please follow these steps:
\begin{enumerate}
    \item Select up to 15 common features relevant to the product group. These features must include product ID, product title, brand, and price. Choose additional features based on their commonality and necessity within the product group.
    \item For each product in the product group, verify the feature values for each selected feature.
\end{enumerate}

Finally, the product tables are reviewed and refined by the authors.

\paragraph{Task 2: QA collection} Annotators are required to fill out a table as shown in Table \ref{qa_annotation_table}. Each row contains a triplet consisting of a \emph{question}, a \emph{long answer}, and a \emph{short answer}, all generated by GPT-4. Annotators should fill the following columns: \emph{Is question reasonable}, \emph{Is long answer correct}, \emph{Refined long answer}, \emph{Is short answer correct} and \emph{Refined short answer}. Please follow these steps:
\begin{enumerate}
    \item \textbf{Evaluate the question}: Verify if the \emph{question} resembles a typical query found in real-world product conversations in online shopping. Select `yes' or `no' in the \emph{Is question reasonable} column. Any question containing harmful information is considered unreasonable and should be labeled as `no'. If `no' is selected, the row will be dropped, and you do not need to proceed with the subsequent steps for that row.
    \item \textbf{Assess the long answer}: Check if the \emph{long answer} correctly responds to the \emph{question}. Select `yes', `no' or `I do not know' in the \emph{Is long answer correct} column. Consider the following special cases:
    \begin{itemize}
        \item If the long answer is ambiguous (e.g., `The product is designed to be waterproof, while some users do not think so.'), mark it as incorrect.
        \item For numerical questions, an answer is considered correct if it fits the real-world scenario and the conclusion is clear. Specific values or ranges (e.g., 5cm, 5cm-10cm, several months) are acceptable if they correspond to the real-world scenario.
        \item If the long answer contains a specific piece of knowledge, verify its accuracy. 
        \item If the long answer is incorrect or does not address the question, and you do not know the correct answer (even after checking the product information table, looking up the product URL, and searching online), select `I do not know'.
        \item Any long answer containing harmful information should be labeled as `I do not know'.
    \end{itemize}
    If you select `I do not know', the row will be dropped, and you do not need to perform the subsequent steps for that row.
    \item \textbf{Refine the long answer}: If you select `no' in step 2, provide a correct long answer in the \emph{Refined long answer} column.
    \item \textbf{Assess the short answer}: Determine whether the \emph{short answer} is correct. A short answer must be `yes', `no', or an entity. Choose `yes' or `no' in the \emph{Is short answer correct} column. Consider the following special cases:
    \begin{itemize}
        \item If the question is a choice and the short answer is `yes' or `no', it is incorrect.
        \item If the question pertains to degrees (e.g. `How durable ... ?') and the short answer is `yes' or `no', it is incorrect.
        \item If the short answer does not align with the long answer, it is incorrect.
    \end{itemize}
    \item \textbf{Refine the short answer}: If you select `no' in step 4, provide a correct short answer in the \emph{Refined short answer} column.
\end{enumerate}

The authors will review the annotation in batches. Specifically, 5\% of each batch will be checked. If the accuracy rate of the checked annotation is below 98\%, the entire batch will be relabeled.

\begin{table}[htbp]
\centering
\caption{An example of the ProductQA annotation table.}
\scalebox{0.55}{
\begin{tabular}{m{4.7cm}m{1.4cm}m{5cm}m{1.6cm}m{1.4cm}m{1.6cm}m{1.7cm}m{1.5cm}}
\toprule[1pt]
\multicolumn{1}{c}{\multirow{2}{*}[-0.2ex]{\textbf{Question}}} & \multicolumn{1}{c}{\textbf{Is question}} & \multicolumn{1}{c}{\multirow{2}{*}[-0.2ex]{\textbf{Long answer}}} & \multicolumn{1}{c}{\textbf{Is long answer}} & \multicolumn{1}{c}{\textbf{Refined}} & \multicolumn{1}{c}{\multirow{2}{*}[-0.2ex]{\textbf{Short answer}}} & \multicolumn{1}{c}{\textbf{Is short answer}} & \multicolumn{1}{c}{\textbf{Refined}} \\
 & \multicolumn{1}{c}{\textbf{reasonable}} & & \multicolumn{1}{c}{\textbf{correct}} & \multicolumn{1}{c}{\textbf{long answer}} & & \multicolumn{1}{c}{\textbf{correct}} & \multicolumn{1}{c}{\textbf{short answer}} \\
\midrule[0.5pt]
What is the size of the neodymium driver used in the JVC HA-EB75 headphones? & [To fill] & The JVC HA-EB75 headphones contain a 13.5 mm neodymium driver in each earpiece, which contributes to the enhanced sound quality. & [To fill] & [To fill] & 13.5 mm & [To fill] & [To fill] \\
\bottomrule[1pt]
\end{tabular}
}
\label{qa_annotation_table}
\end{table}

\section{Broader impact}\label{sec:Broader impact}

\subsection{Positive broader impact}

(1) We created ProductQA, a dataset of 88,229 QA pairs across 26 product groups. This dataset provides a comprehensive evaluation environment for LLM agents, addressing real-world challenges such as managing historical information and accumulated knowledge, using tools, interacting with humans, performing self-evaluation, conducting reflection, and adapting to new tasks. We believe that ProductQA can advance the research in LLM agents.

(2) \agentnm serves as a general framework that supports a wide range of extensions. Agents within the framework can use more tools, perform complex reasoning using LLMs alone or in combination with other tools, and self-improve by accumulating experiences and knowledge. \agentnm provides an architecture for creating powerful agents with the potential to achieve human-level intelligence.

(3) \agentnm supports proactive seeking advice from human experts, ensuring a high level of accuracy for applications, even when dealing with challenging questions. Within this framework, we can manage the trade-off between accuracy and human cost. These features enable \agentnm agents to be applied in real-world scenarios.

\subsection{Negative broader impact}

In practical applications, LLM agents exhibit superior capabilities compared to standalone LLMs. Our research validates that the AGILE framework is a highly effective approach for optimizing LLM agents. However, this improvement also increases the potential risks of harmful applications. Therefore, it is crucial to intensify research on the safety and responsible use of LLM agents.

\section{Limitations} \label{sec:Limitations}

(1) Due to resource constraints, our experiments primarily utilize LLMs with 7B or 13B parameters within \agentnm. We expect that applying \agentnm framework to larger models will result in more powerful agents, especially in planning and reasoning. Expanding \agentnm to larger LLMs is our future work.

(2) Our ProductQA dataset includes QA pairs from 20 product groups in the training set. Due to resource constraints, we randomly selected 6 of the 20 groups for training our \agentnm agent. Despite using a subset of training data, our \texttt{agile-vic13b-ppo} shows significant improvements over GPT-4 agent in accuracy and total score. Future work could enhance the agent's capabilities by training on a larger and more diverse dataset, potentially further improving performance and effectiveness.

\section{Ethical considerations} \label{sec:Ethical Considerations}

ProductQA is constructed based on the Amazon Review Dataset. We only use the review data for each product without any user personal information, such as the identity of the reviewers. 

All data in ProductQA are annotated by human annotators, as described in Appendix~\ref{annotation_guidelines}. Any data containing harmful information is removed during the annotation process. 

The annotation team has 20 annotators, each holding at least a college degree, and employed by a commercial data annotation company. We have contracted this company and paid them for the annotation work at a market price.

\section{Prompt templates} \label{sec:Prompt templates}
\begin{figure}[ht]
\centering
\includegraphics[width=1.0\textwidth]{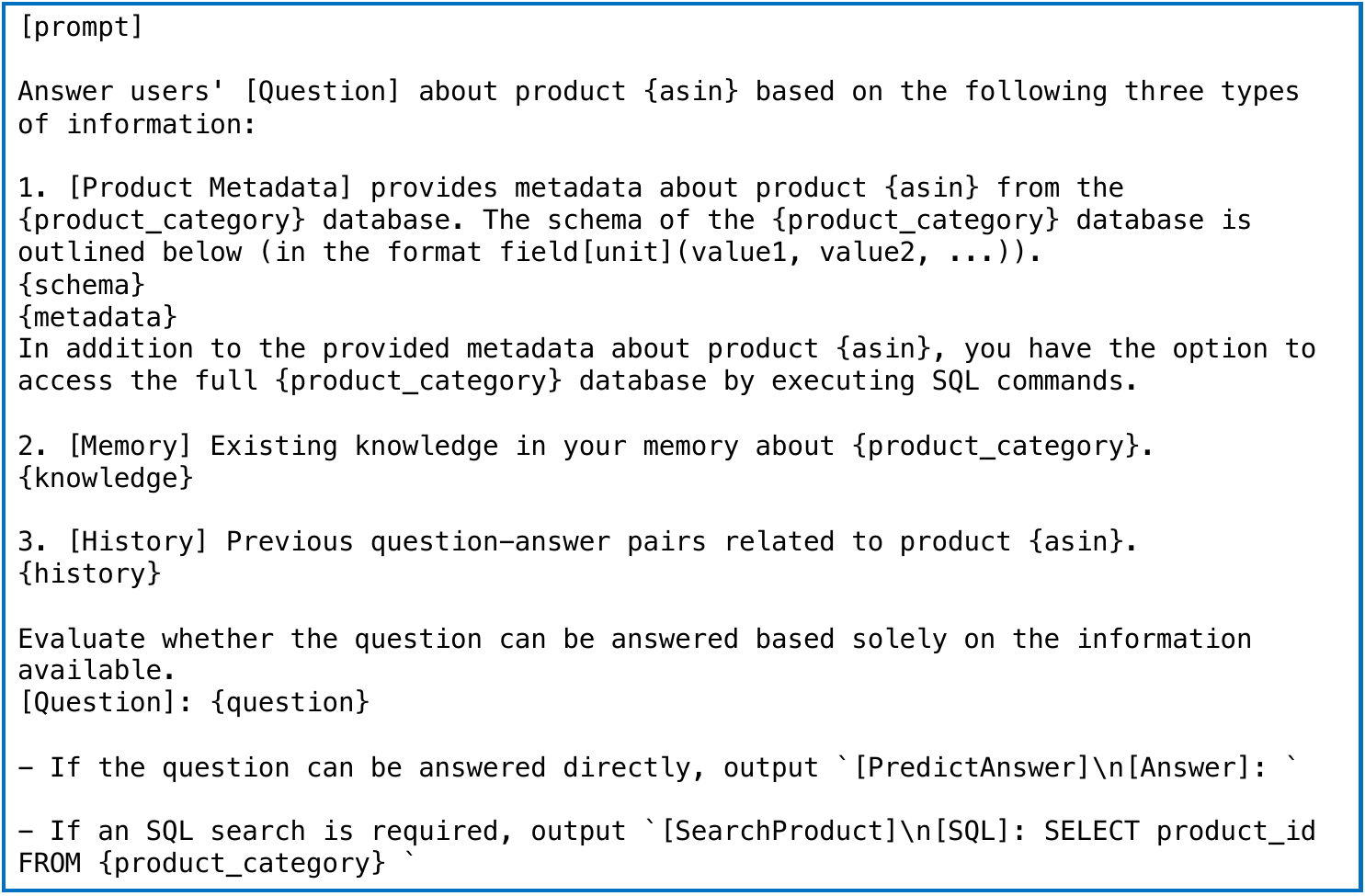}
\caption{The prompt for \texttt{gpt3.5-prompt} and \texttt{gpt4-prompt} on ProductQA.}
\label{fig:gpt_productqa_prompt}
\end{figure}

\begin{figure}[ht]
\centering
\includegraphics[width=1.0\textwidth]{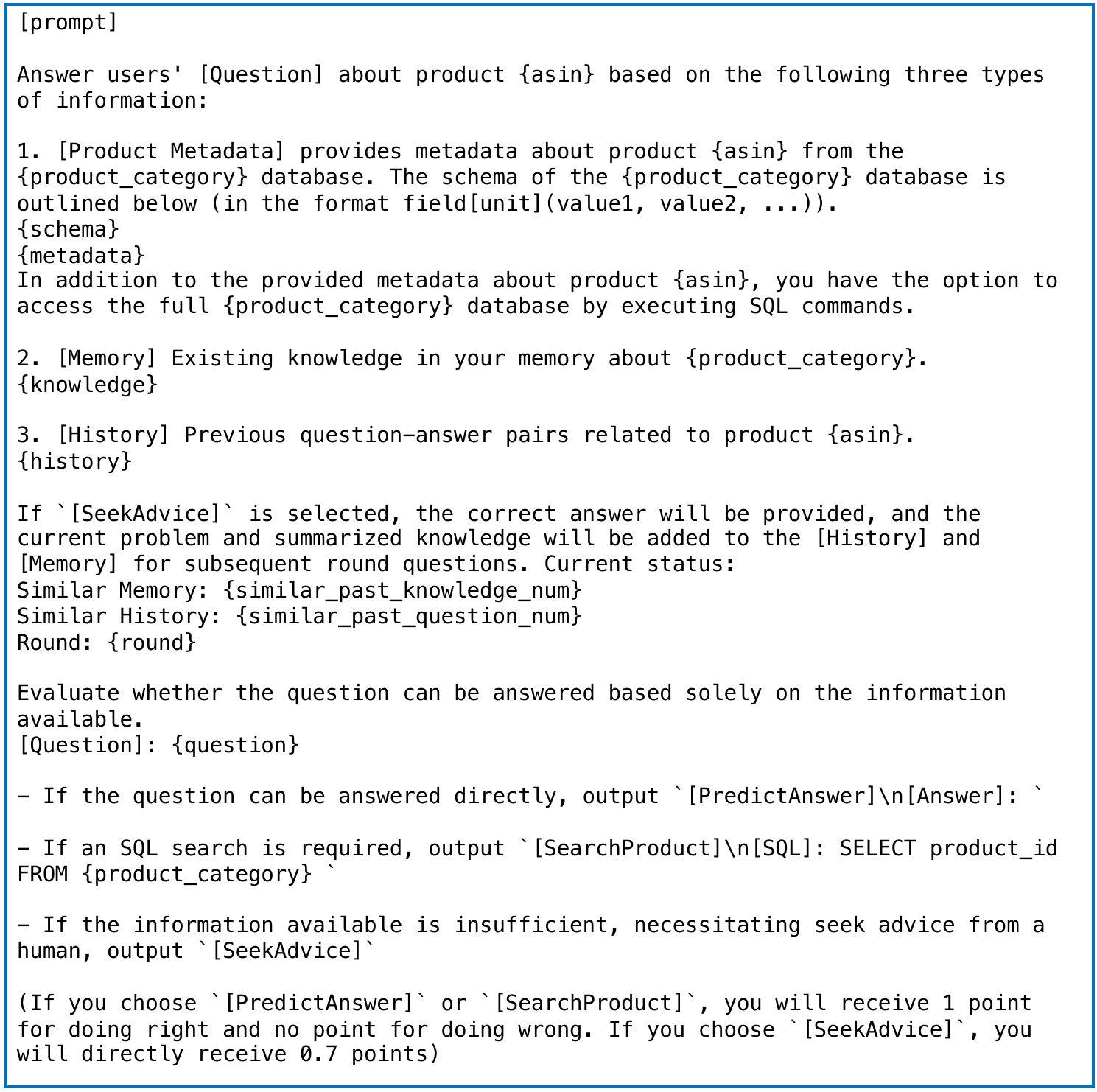}
\caption{The prompt for \texttt{agile-vic13b-prompt}, \texttt{agile-gpt3.5-prompt}, and \texttt{agile-gpt4-prompt} on ProductQA.}
\label{fig:agile_gpt_productqa_prompt}
\end{figure}

\paragraph{Prompt templates for ProductQA}
Figure \ref{fig:gpt_productqa_prompt} shows the prompt template for \texttt{gpt3.5-prompt}, \texttt{gpt4-prompt}. Figure \ref{fig:agile_gpt_productqa_prompt} provides the prompt template for \texttt{agile-vic13b-prompt}, \texttt{agile-gpt3.5-prompt}, and \texttt{agile-gpt4-prompt}.
We leave the "\{knowledge\} and "\{history\}" empty when evaluate \texttt{gpt3.5-prompt} and \texttt{gpt4-prompt}.

\begin{figure}[ht]
\centering
\includegraphics[width=1.0\textwidth]{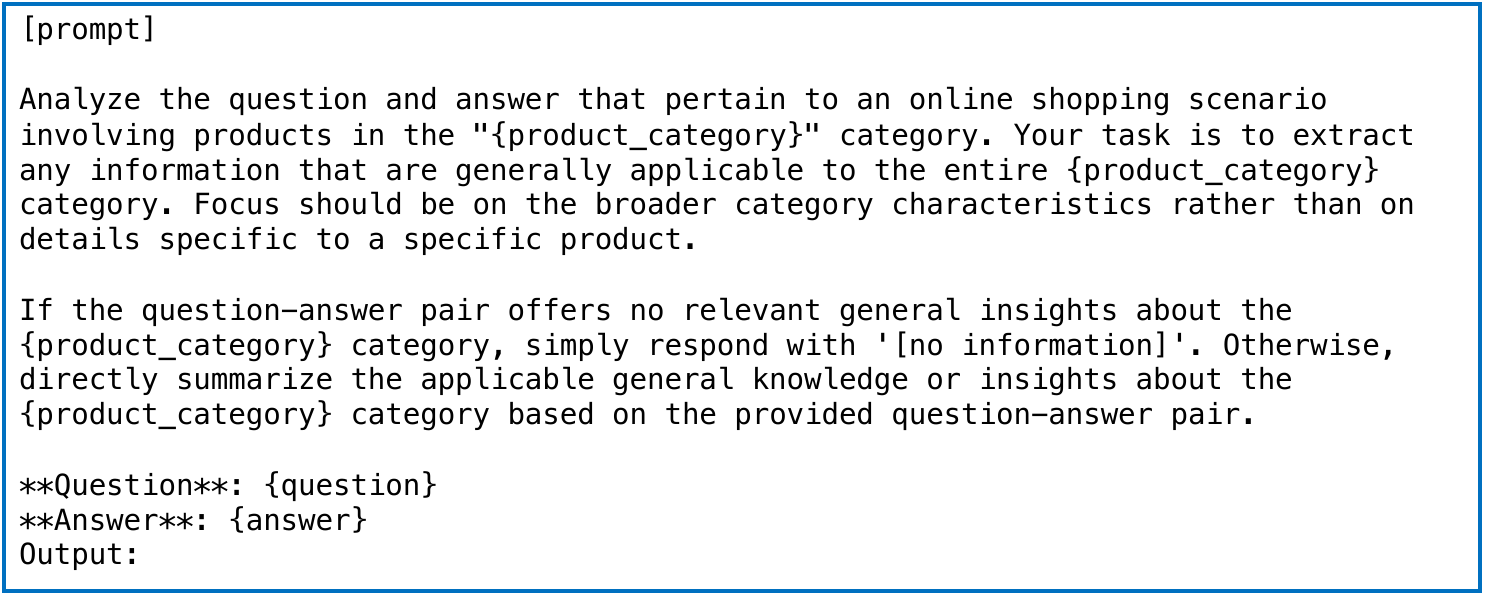}
\caption{The prompt for reflection on ProductQA.}
\label{fig:reflect_productqa_prompt}
\end{figure}

The prompt template for reflection is shown in Figure \ref{fig:reflect_productqa_prompt}.

\begin{figure}[ht]
  \centering
   \includegraphics[width=1.0\textwidth]{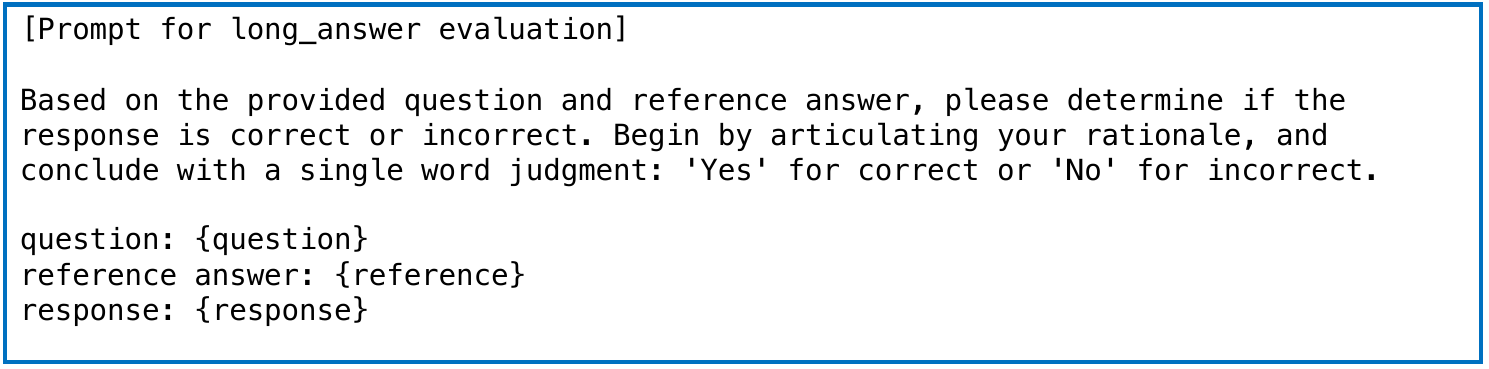}
   \caption{The prompt for long answer evaluation on ProductQA.}
   \label{fig:prompt_long_answer_eval}
\end{figure}

The prompt template for long answer evaluation is shown in Figure~\ref{fig:prompt_long_answer_eval}.

\begin{figure}[ht]
\centering
\includegraphics[width=1.0\textwidth]{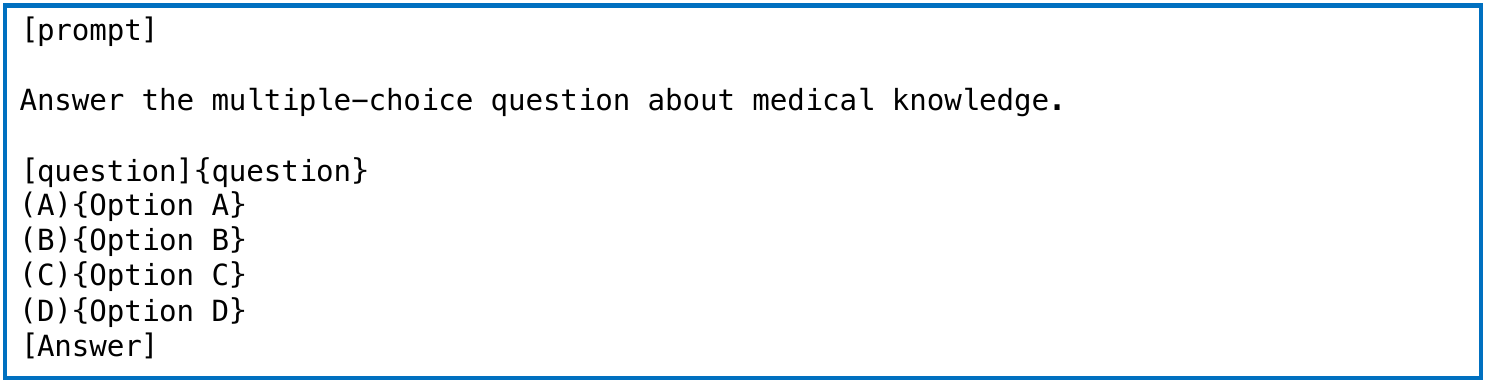}
\caption{The prompt for \texttt{Meerkat-7b-prompt} on MedMCQA.}
\label{fig:gpt_medmcqa_prompt}
\end{figure}

\begin{figure}[ht]
\centering
\includegraphics[width=1.0\textwidth]{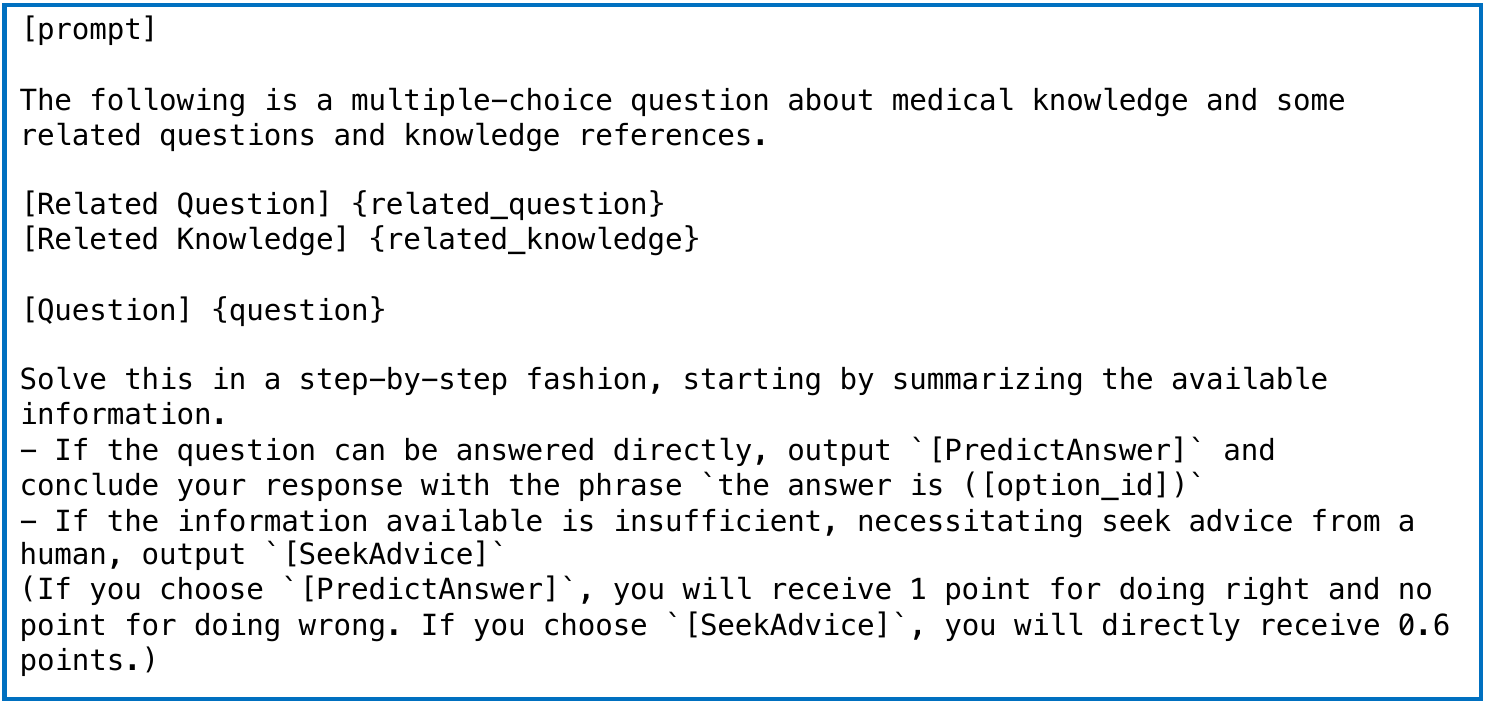}
\caption{The prompt for \texttt{agile-gpt3.5-prompt} and \texttt{agile-gpt4-prompt} on MedMCQA.}
\label{fig:agile_gpt_medmcqa_prompt}
\end{figure}

\begin{figure}[ht]
\centering
\includegraphics[width=1.0\textwidth]{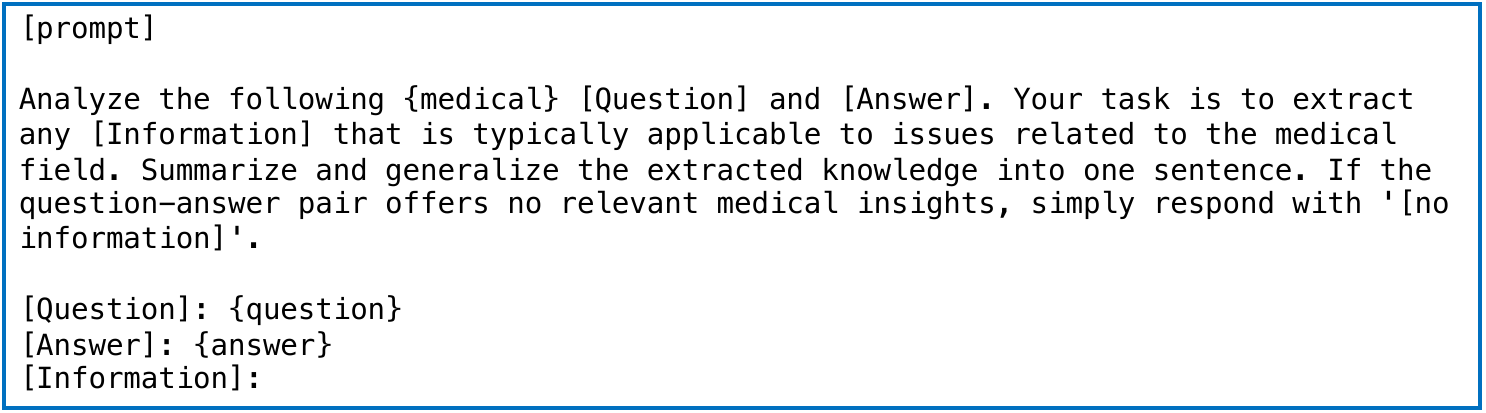}
\caption{The prompt for reflection on MedMCQA.}
\label{fig:reflect_medmcqa_prompt}
\end{figure}

\paragraph{Prompt templates for MedMCQA}
Figure \ref{fig:gpt_medmcqa_prompt} provides the prompt template for \texttt{Meerkat-7b-prompt}. Figure \ref{fig:agile_gpt_medmcqa_prompt} illustrates the prompt template for \texttt{agile-gpt3.5-prompt}, \texttt{agile-gpt4-prompt}. We leave the "\{related\_question\} and "\{related\_knowledge\}" empty when evaluate \texttt{gpt3.5-prompt} and \texttt{gpt4-prompt}. The prompt template for reflection is shown in Figure \ref{fig:reflect_medmcqa_prompt}.

\begin{figure}[ht]
\centering
\includegraphics[width=1.0\textwidth]{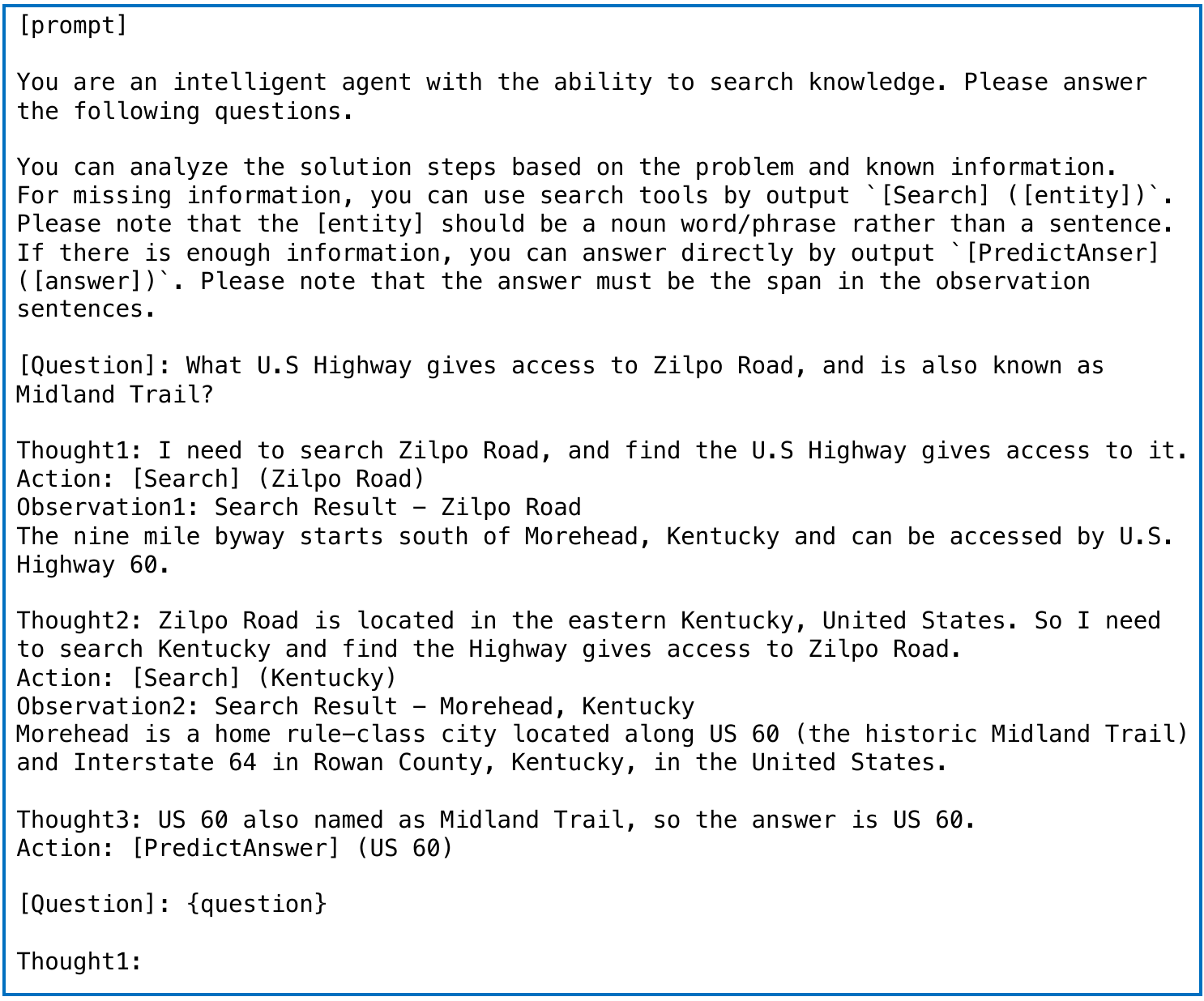}
\caption{The prompt for \texttt{ReAct-gpt4-prompt} on HotPotQA.}
\label{fig:agile_react_hotpotqa_prompt}
\end{figure}

\begin{figure}[ht]
\centering
\includegraphics[width=1.0\textwidth]{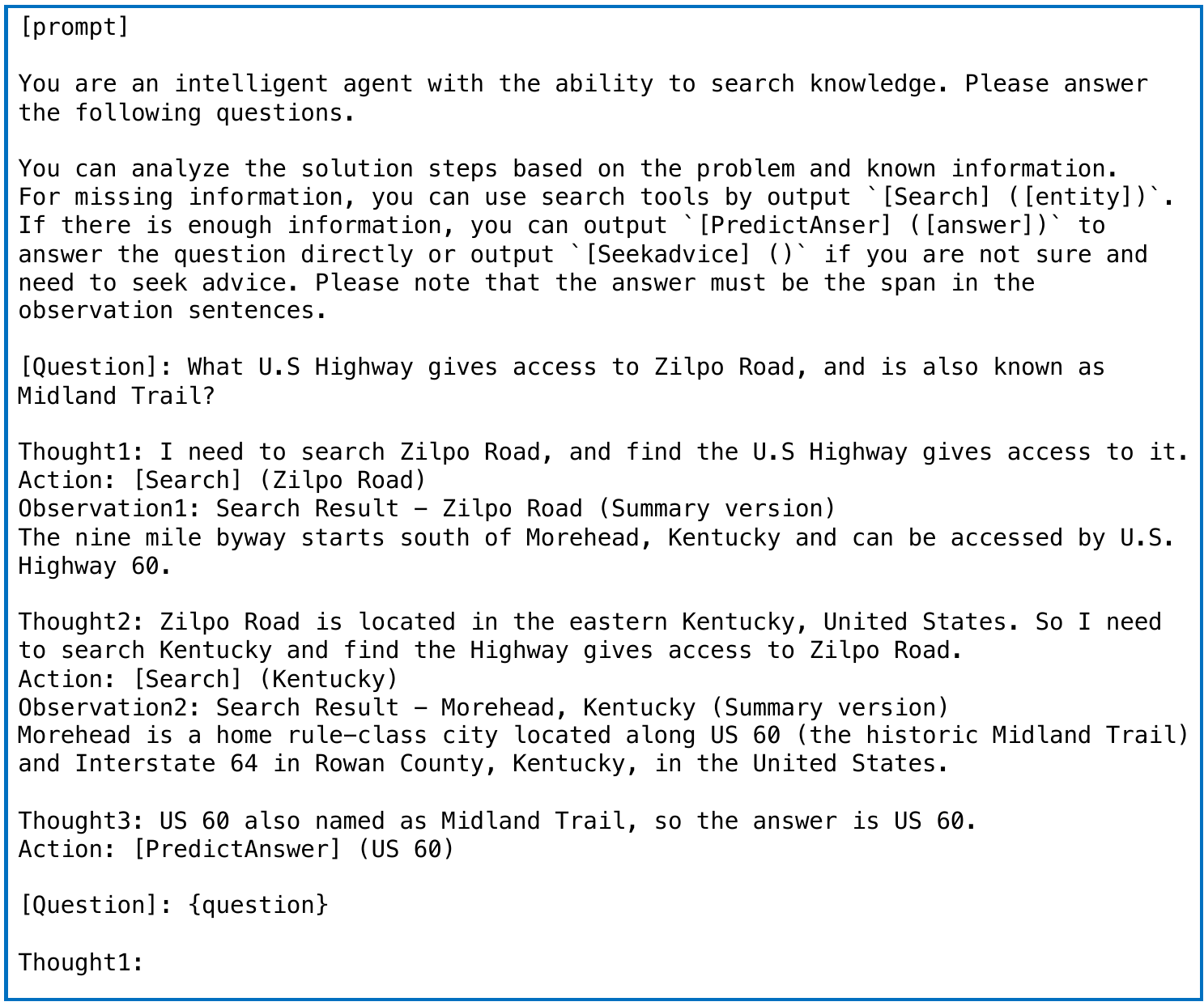}
\caption{The prompt for \texttt{agile-gpt4-prompt} on HotPotQA.}
\label{fig:agile_gpt_hoppotqa_prompt}
\end{figure}

\begin{figure}[ht]
  \centering
   \includegraphics[width=1.0\textwidth]{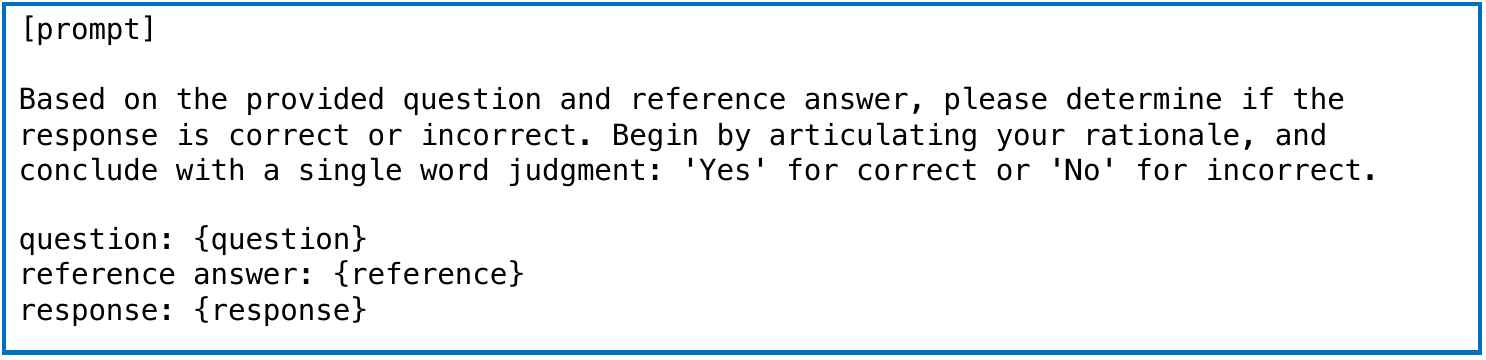}
   \caption{The prompt for answer evaluation on HotPotQA.}
   \label{fig:prompt_answer_eval_hotpotqa}
\end{figure}

\paragraph{Prompt templates for HotPotQA}
Figure \ref{fig:agile_react_hotpotqa_prompt} provides the prompt template for \texttt{ReAct-gpt4-prompt}. Figure \ref{fig:agile_gpt_hoppotqa_prompt} illustrates the prompt template for \texttt{agile-gpt4-prompt}. The prompt template for answer evaluation is shown in Figure~\ref{fig:prompt_answer_eval_hotpotqa}.

\end{document}